\documentclass{article}

\usepackage{PRIMEarxiv}

\usepackage[utf8]{inputenc} % allow utf-8 input
\usepackage[T1]{fontenc}    % use 8-bit T1 fonts
\usepackage{hyperref}       % hyperlinks
\usepackage{url}            % simple URL typesetting
\usepackage{booktabs}       % professional-quality tables
\usepackage{amsfonts}       % blackboard math symbols
\usepackage{nicefrac}       % compact symbols for 1/2, etc.
\usepackage{microtype}      % microtypography
\usepackage{lipsum}
\usepackage{fancyhdr}       % header
\usepackage{graphicx}       % graphics
\usepackage{amsmath}            % Para las referencias a ecuaciones con \eqref
\usepackage{epstopdf}           % Para poder insertar figuras .eps al compilar con PDFLATEX
\usepackage{flushend}           % Para igualar las columnas de la altima pagina
\usepackage{algorithm}
\usepackage{algpseudocode}
\usepackage{hyperref}
\usepackage{authblk}
\graphicspath{{media/}}     % organize your images and other figures under media/ folder

\algnewcommand\algorithmicforeach{\textbf{For Each}}
\algdef{S}[FOR]{ForEach}[1]{\algorithmicforeach\ #1\ \algorithmicdo}       
\title{From Single Aircraft to Communities: A Neutral Interpretation of Air Traffic Complexity Dynamics
}

\author[1, *]{Ralvi Isufaj}
\author[1]{Marsel Omeri}
\author[1]{Miquel Angel Piera}
\author[1]{Jaume Saez Valls}
\author[2]{Christian Eduardo Verdonk Gallego}
\affil[1]{Logistics and Aeronautics Group, Autonomous University of Barcelona, Sabadell, Spain}
\affil[2]{CRIDA A.I.E, Madrid, Spain}
\affil[*]{Corresponding author: ralvi.isufaj@uab.cat}

\begin{document}
\maketitle

\begin{abstract}
Present air traffic complexity metrics are defined considering the interests of different management layers of ATM. These layers have different objectives which in practice compete to maximize their own goals, which leads to fragmented decision making. This fragmentation together with competing KPAs requires transparent and neutral air traffic information to pave the way for an explainable set of actions. In this paper, we introduce the concept of single aircraft complexity, to determine the contribution of each aircraft to the overall complexity of air traffic. Furthermore, we describe a methodology extending this concept to define complex communities, which are groups of interdependent aircraft that contribute the majority of the complexity in a certain airspace. In order to showcase the methodology, a tool that visualizes different outputs of the algorithm is developed. Through use-cases based on synthetic and real historical traffic, we first show that the algorithm can serve to formalize controller decisions as well as guide controllers to better decisions. Further, we investigate how the provided information can be used to increase transparency of the decision makers towards different airspace users, which serves also to increase fairness and equity. Lastly, a sensitivity analysis is conducted in order to systematically analyse how each input affects the methodology.
\end{abstract}

% keywords can be removed

\section{Introduction}
Air Traffic Management (ATM) is a complex socio-technical system comprised by three main layers; air space management (ASM), air traffic flow management (ATFM) and air traffic control (ATC) \cite{nieto2020collision}, whose performance is measured through various Key Performance Areas (KPAs), of which some of the most important are safety, capacity, cost-efficiency and environment \cite{days2014modelling}. Although part of ATM structure, each of these layers have different objectives which in practice compete to maximize their own goals. Several authors point out that complex interdependencies among the decision layers can cause unnecessary penalization on some KPAs to improve others.
In \cite{ruiz2018novel} authors claim as a result of early SESAR projects such as STREAM \cite{ranieri2011stream} that the exact relationship among these KPAs is still not well understood and should be further studied in future research. Moreover, recent finalized projects such as APACHE \cite{netjasov2017assessment} targeting the analysis of the interdependencies between the different KPAs by capturing the Pareto-front of ATM, still claims that further research is needed to uncover the inter-dependencies between the different KPA’s \cite{netjasov2018assessment}. An important challenge that must be overcome to reach such a Pareto-front of ATM KPAs is the lack of an effective coordination mechanism (i.e., system behaviour) among ATM subsystems and corresponding Decision Support Systems (DSS).

Figure 1 illustrates a conceptual framework for ATM-related quality of services based in \cite{Canso,ATM_performance_USEU}. The idea is to visualize all the components (i.e., subsystem, KPA, DSS) and mark the inter-relations between them. The distinction of components is done by color-coding based on their functions and definitions. Each component is connected to one or other components. The component that is attached to the arrowhead indicates that it is the one which is affected by the other. Note that connectors have different colours, based on the origin of the subsystems (e.g., ATC). According to \cite{battistella2013methodology}, the represented interactions can be considered as complex adaptive systems (CAS), which typically include multiple loops and multiple feedback paths between many interacting entities, as well as inhibitory connections and preferential reactions. 

Furthermore, an illustrative example highlights the interactions among the DSS that target the airspace management (AM) and the trajectory management (TM) and capacity management. This is a common example to highlight how these decisions can cause a chain of reactions resulting in en-route ATC inefficiency, delays at airport (taking-off and landing) and en-route ATFM inefficiency \cite{gulding2010us}. To date, the lack of a formal analysis between these control mechanisms leads to a lack of transparency and coordination between different DSS that could improve ATM performance.

\begin{figure*}[h]
 \centering
  \includegraphics[scale=0.3]{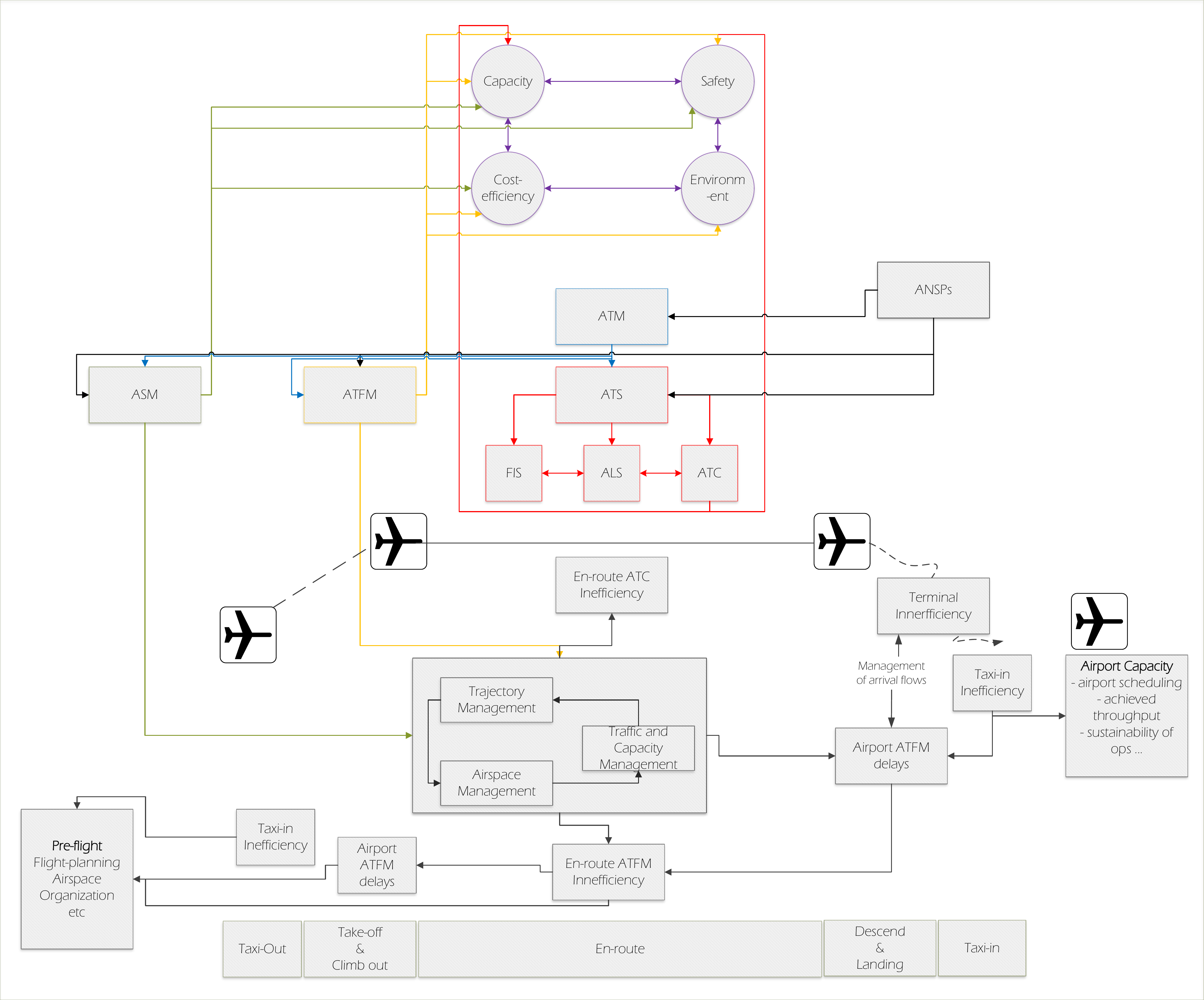}
  \caption{Conceptual Framework for ATM-related quality of services.}
\end{figure*}
Complex interdependencies among the mentioned KPAs have been key for the aeronautic community to accept the confusing term \textit{emergent dynamics} which in fact is justified by the un-modelled behavioural dynamics among these objectives (i.e., capacity, safety, and efficiency). For instance, consequences of a small reduction of a sector capacity (considering ground weather conditions) usually is tackled by over-conservative ATM actions to avoid safety issues at a cost of penalizing flight efficiency. In the other hand, other solutions with better information (weather information at flight deck) could improve flight efficiency. In \cite{bedau2008downward} a difference between \textit{weak emergence dynamics} and \textit{strong emerge dynamics} is introduced to differentiate between  micro-level interactions among subsystems (weak) and emergence caused by irreducible macro causal mechanisms (strong). Authors of this paper, accept that unpredictable decision-making processes carried by a human actor (i.e. aircraft pilot, air traffic controller) justify the term of \textit{strong emergent dynamics} in ATM which can be observed in several ATM socio-technical subsystems \cite{piera2022socio}. 

To avoid an ATM system dynamics ruled by \textit{strong emergent dynamics} due to abrupt human behaviour, in this paper a new methodological framework to enhance a common understanding among the different stakeholders is proposed. Worthwhile to highlight that as a result of the implemented methodology, the framework allows ANSP work closely with the rest of stakeholders avoiding over constraining solutions. Moreover, the proposed framework paves the way for a shared situational awareness in which the effects of unpredictable decision-making processes carried by human actors is mitigated by the consensus reached among the different actors, transforming the \textit{ strong emergent dynamics} into \textit{weak emergent dynamics} of ATM.

The core idea behind the proposed methodology, lies in understanding complexity evolution of the overall air traffic. In this paper we extend the complexity notion and indicators proposed in \cite{isufaj2021spatiotemporal}, by introducing single aircraft complexity concept. We define single aircraft complexity as the contribution that individual aircraft have to the overall sector complexity. Furthermore, we identify groups of interdependent aircraft that form high complex spatio-temporal areas. This method identifies the individual and community (i.e., groups of aircraft) contributions in an on-line fashion and gives information about the creation, evolution and disappearance of communities. This information could enhance equity, fairness at aircraft or airline granularity level together with NM and ANSP service performance. The proposed method could make an immediate impact in the smooth transition between ATM layers and DSS tools. In other words, each subsystem should comply with their operational performance specifications (e.g., ATC must prevent loss of separations), while decreasing mutual penalization.

The rest of this work is structured as follows: In Section 2, we describe in detail the methodology. Section 3 provides an overview of the experimental setup. In Section 4, we present and discuss several use cases based on synthetic and real traffic, as well as an extensive sensitivity analysis. We draw conclusions and discuss future steps in Section 5.

\section{Methodology}
\subsection{Spatiotemporal Graph-based Complexity Indicators}
While there have been many different definitions for airspace complexity, in this work we will extend the one introduced in \cite{isufaj2021spatiotemporal}. There, the authors focus on defining complexity for a certain volume of space (e.g., a sector) during a time window of interest and model air traffic as a dynamic graph $G(t) = (V(t),E(t))$. The set of vertices $V(t)$ for time $t$ is comprised of the aircraft present in the sector at the time, while the set of edges $E(t)$ are the interdependencies between each pair aircraft at time $t$. Interdependencies are defined based on the distance between aircraft, more specifically, if two aircraft are closer than a certain threshold then there will be a weight between these two aircraft. The closer the aircraft are, the bigger will be the weight of the edge between these two aircraft, which means that the graph is \textbf{weighted} and \textbf{undirected}. Weights are normalized to be between 0 and 1. 

In their work, they are interested for en-route traffic at the tactical level, therefore they define the weight of the edge to be maximal (i.e., 1) when there is a loss of separation between a pair of aircraft (5 NM horizontally and 1000 feet vertically). Horizontal and vertical interdepndencies are calculated separately and the overall interdependency between two aircraft is the average of the two. Formally, this is defined as:
\begin{equation}
wh_{i,j}(t) = \begin{cases}
 1 \text{ if } dh_{i,j}(t) \leq H \\
 0 \text{ if } dh_{i,j}(t) \geq thresh_h \\
 \frac{thresh_h - dh_{i,j}(t)}{thresh_h - min_h} \text{ otherwise}
\end{cases}
\end{equation}
\begin{equation}
wv_{i,j}(t) = \begin{cases}
 1 \text{ if } dv_{i,j}(t) \leq V \\
 0 \text{ if } dv_{i,j}(t) \geq thresh_v \\
 \frac{thresh_v - dv_{i,j}(t)}{thresh_v - min_v} \text{ otherwise}
\end{cases}
\end{equation}
\begin{equation}
w_{i,j}(t) = \begin{cases}
\frac{wh_{i,j}(t) + wv_{i,j}(t)}{2} \text{ if } wh_{i,j}(t) > 0 \And wv_{i,j}(t) > 0 \\
0 \text{ otherwise}
\end{cases}
\end{equation}
where $wh_{i,j}(t)$ and $wv_{i,j}(t)$ are the horizontal and vertical weights at time $t$. Furthermore, $dh_{i,j}(t)$ and $dv_{i,j}(t)$ are the distances, $H$ and $V$ are the safety distances and $thresh_h$ and $thresh_v$ are the thresholds. 

Airspace complexity is treated as a multifaceted notion and the authors propose four indicators that quantify topological information and combine it with the severity of the interdependencies. We will briefly describe these indiators, however we refer the reader to \cite{isufaj2021spatiotemporal} for a detailed overview.

\textit{Edge Density} (ED) measure how many edge the graph has compared to the number of edges in a fully connected graph of the same size with maximal edge. Formally:
\begin{equation}
ED(G,t) = \frac{\sum_{(i,j) \in E} {w_{i,j}(t)}}{A(V_t)}, A(V_t) = \frac{|V_t|(|V_t| - 1)}{2}
\end{equation}
where $|V_t|$ is the number of vertices in the graph at time t and $A(V_t)$ is the maximal number of edges. 

\textit{Strength} measures the severity of pairwise interdependencies. It is obtained by extending the definition of vertex degree to account for edge weights:
\begin{equation}
s(i,t) = \sum_{j=1}^N{w_{i,j}(t)}
\end{equation}
\textit{Clustering Coefficient} (CC) measures the local cohesiveness and gives information regarding the local neighbourhood of each vertex (i.e., aircraft). Formally, it is calculated as follows:
\begin{equation}
CC(i,t) = \frac{\sum_{j,k}{(w_{i,j}(t)+w_{j,k}(t))}}{ 2 \cdot (s(i,t) (deg(i,t) -1)}, \forall (i,j,k) \in \mathcal{T}(t)
\end{equation}
\textit{Nearest Neighbor Degree} (NND) calculates a local weighted average degree of the nearest neighbour for each aircraft:
\begin{equation}
NND(i,t) = \frac{\sum_{j=1}^N {w_{i,j}(t)deg(j,t)}}{s(i,t)}
\end{equation}
The first indicator is inherently a global measure, while the remaining three indicators are turned into global measures by taking the average across vertices in the graph.

The overall complexity of the sector is chosen to be given as the evolution in time of each indicator and the authors argue that this results in a more nuanced overview of complexity.

\subsection{Single Aircraft Complexity}
While the previously described methodology gives a more nuanced view of complexity than simpler metrics (e.g., dynamic density \cite{laudeman1998dynamic}), it still suffers from a common drawback of the majority of existing complexity metrics: a lack of interpretability of the complexity scores. More specifically, given a certain traffic configuration, existing methods cannot provide information as to which areas of the sector are causing most of the complexity and how much of it they are causing. Furthermore, \cite{isufaj2021spatiotemporal} do not discuss ways how to combine the information provided by each indicator. 

In this work, we exploit an inherent characteristic of complexity defined based on graph theory to overcome this major drawback. As previously mentioned, three of the four indicators described, which will be in the focus of our work, can be defined in terms of single aircraft, with the overall being the average across aircraft. This means that we can generate a complexity score for every single aircraft present in the sector in time $t$ and for every indicator without loss of information. However, this is not sufficient, as single scores would simply induce an order between aircraft for every indicator without providing any information regarding the overall situation of complexity in the sector. Another issue with this method is that it contains redundant information, as interdependencies are undirected, e.g., 
$A \rightarrow B$ and $B \rightarrow A$. Finally, it is not clear how to interpret the scale and value of the complexity scores. Furthermore, since the indicators have a different range of possible values that they can take, it is not trivial how to combine the individual absolute scores of every aircraft. 

\begin{figure}[h]
\centering
\includegraphics[scale=0.5]{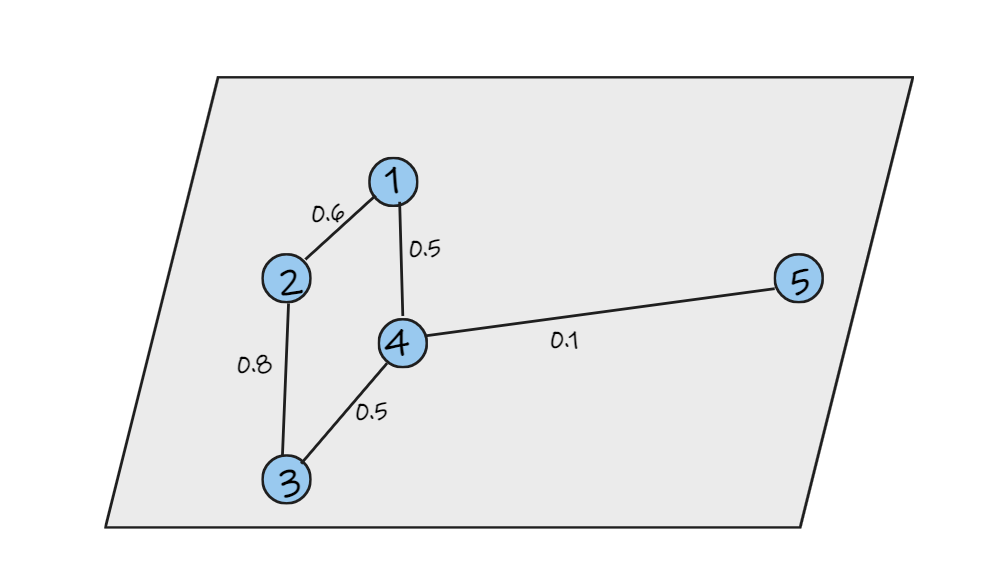}
\caption{Example of an arbitrary sector.}
\label{fig:ex1}
\end{figure}

In this paper, we propose to slightly change perspective and calculate the contribution of each aircraft to the overall sector complexity. This results not in an absolute score, but in a percentage that is relative to what is currently happening in the sector at the time. Without loss of generality, we will show how the contribution is calculated for the \textit{strength} indicator. Let us consider an arbitrary sector which at time $t$ is occupied by the aircraft shown in Figure \ref{fig:ex1}. Following Equation 5, we can determine the individual strengths of the aircraft, shown in Table \ref{tbl:str}. 

The individual contributions are calculated as which part of the whole strength each aircraft is responsible for. Differently from the original definition of the overall strength (the average), in this work we slightly modify this definition and take the whole as the sum of all aircraft. In this case, the overall strength would be:
\begin{equation}
    s(t) = \sum_{i=1}^N s(i,t)
\end{equation}
The contribution of aircraft $i$ to the overall strength would then be:
\begin{equation}
    c_s(i,t) = \frac{s(i,t)}{s(t)}
\end{equation}
Using this formula, we can now determine the contribution of each aircraft to the overall strength as shown in Table \ref{tbl:str}. 
Following a similar method we can generalize how to calculate the contribution of an aircraft for every complexity indicator relevant to our work (strength, CC, NND):
\begin{equation}
    c_\mathcal{I}(i,t) = \frac{\mathcal{I}(i,t)}{\mathcal{I}(t)}      \forall \mathcal{I} \in [strength, CC, NND]
\end{equation}

\begin{table}[]
\centering
\begin{tabular}{@{}lll@{}}
\toprule
Indicator & Value & Percentage \\ \midrule
$s_1$      & 1.1   & 22\%       \\ 
$s_2$      & 1.4   & 28\%       \\ 
$s_3$      & 1.3   & 26\%       \\ 
$s_4$      & 1.1   & 22\%       \\ 
$s_5$      & 0.1   & 2\%        \\ \bottomrule
\end{tabular}
\caption{Individual contributions for the strength indicator.}
\label{tbl:str}
\end{table}
This method allows us to meaningfully combine the contributions for all three complexity indicators. As we are considering the contributions relative to the current situation in the sector, we can observe what percentage of the overall sector complexity each aircraft is responsible for. While there are different ways this can be achieved depending on the use case, in this work we choose the average of non-zero valued indicators at the sector. Only the indicators that have a non-zero value are used, as we are interested in showing the contribution to the existing complexity in the sector as quantified by the indicators. For instance, if $CC(t) = 0$ then this indicator is not a source of complexity for the sector at time $t$, therefore aircraft should not be attributed with contributing to this indicator. This is formalised as:
\begin{equation}
    c(i,t) = \frac{\sum_{\mathcal{I} \in [strength, CC, NND] }c_\mathcal{I}(i,t)}{|\{\mathcal{I} \in [strength, CC, NND] : \mathcal{I} > 0\}|} \times 100
\end{equation}
where the denominator is the cardinality of the set of non-zero valued indicators at time $t$. 

\begin{figure}[h]
\centering
\includegraphics[width=1.0\linewidth]{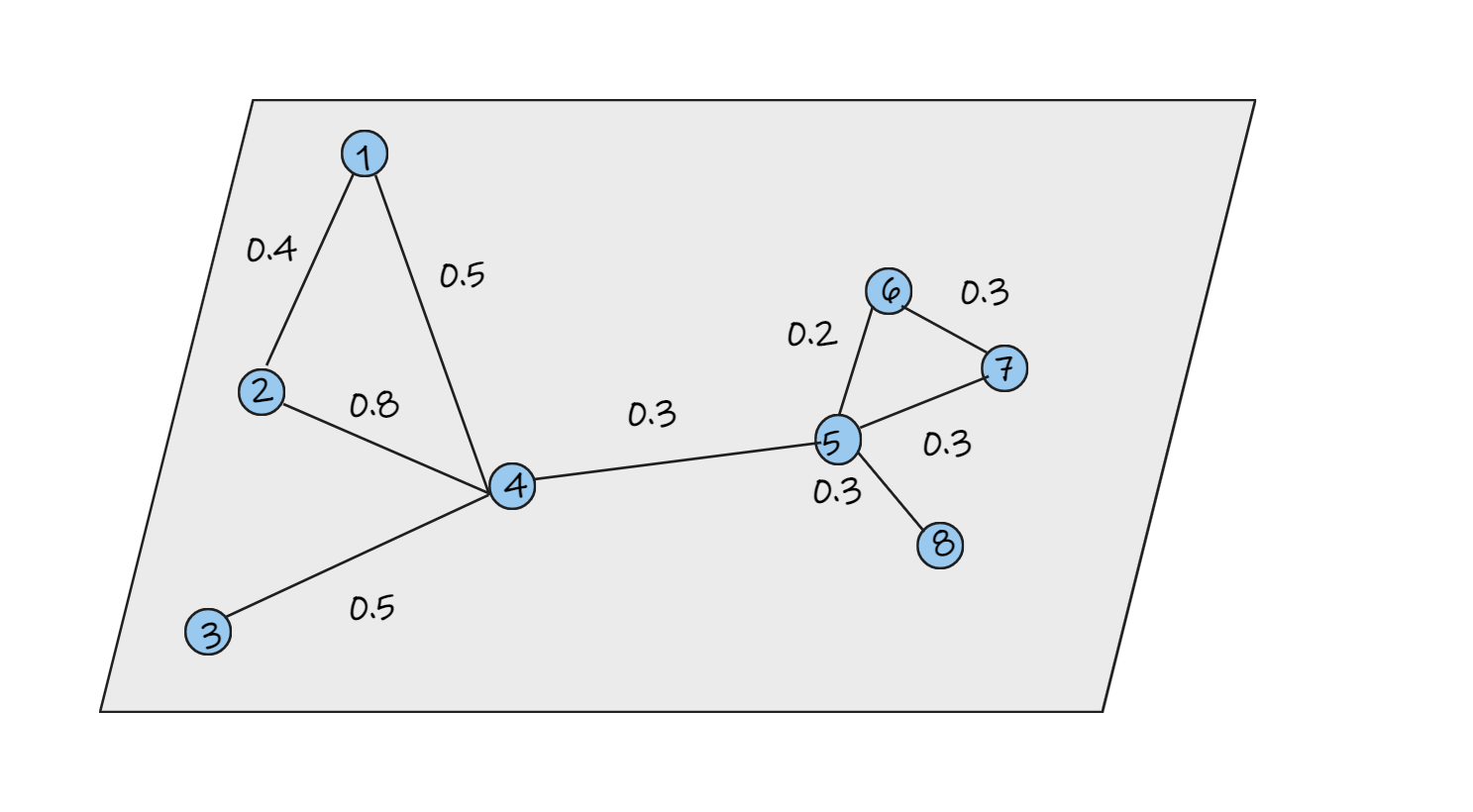}
\caption{More complex example.}
\label{fig:ex2}
\end{figure}

This methodology is illustrated with the example in Figure \ref{fig:ex2}, and the results in Table \ref{tbl:ex2}. As it can be seen, aircraft 2 contributes the most to the overall complexity with 20.45\%. This example shows that the methodology successfully combines the information of all complexity indicators, as aircraft 2 is part of a cluster with 1 and 4 and also has a strong interdependency with aircraft 4 which increases its strength and NND scores. The topology of the subgraph comprised of aircraft 1-4 is mirrored from the subgraph formed of aircraft 5-8. However, we note that the interdependencies in the latter are weaker, which is correctly reflected in the contributions of these aircraft. 

Nevertheless, we note again that the method of combining the contributions from each complexity indicator can depend on the use-case, using a weighted average instead. For instance, let us assume that this methodology will be used by the ATC. In that case, it could be reasonable that the most important indicator is strength, as it directly informs about the distance of the aircraft and how close they are to a loss of separation. Such information could be provided by weighing more the contributions from the strength indicator, while maintaining some of the information provided by the other two indicators. 
\begin{table}[]
\centering
\begin{tabular}{@{}ll@{}}
\toprule
Aircraft & Contribution \\ \midrule
1        & 18.76\%      \\
2        & 20.45\%      \\
3        & 11.61\%      \\
4        & 14.23\%      \\
5        & 9.01\%      \\
6        & 11.12\%      \\
7        & 11.85\%      \\
8        & 10.22\%      \\ \bottomrule
\end{tabular}
\label{tbl:ex2}
\caption{Contribution to complexity from each aircraft in Figure \ref{fig:ex2}}
\end{table}

\subsection{Detection of Complex Spatiotemporal Communities}
Using the concept of single aircraft complexity introduced in the previous section, it is possible to find groups of interdependent aircraft that contribute the majority of the complexity in the sector. Finding tightly interdependent groups of nodes in a graph (i.e., communities) is a well known problem in graph theory, with many existing algorithms providing high quality communities efficiently, such as the Louvain and Leiden algorithms \cite{blondel2008fast, traag2019louvain}. However, these algorithms optimize for \textit{modularity}, which is a quantity that measures the density of connections within a community. Graphs with a high modularity score will have many connections within a community but only few pointing outwards to other communities. Briefly, the algorithms explore for every node if its modularity score might increase if it changes its community to one of its neighboring nodes. 

Given the definition of modularity, it could be reasonable to expect that the communities that these algorithms find could coincide to high complexity communities, but the verification of this assumption will simply increase the runtime of the algorithm. Nevertheless, the more important issue of the aforementioned algorithms is that communities tend to share at least some edges between them. For the application in this paper, it is unclear how this situation ought to be treated. 

To illustrate this point, let us take the graph in Figure \ref{fig:ex2}. In this case, the Louvain and Leiden algorithms output two communities: one comprised of aircraft $\{1,2,3,4\}$ and the other comprised of aircraft $\{5,6,7,8\}$. However, these communities are connected, as there is an edge between 4 and 5. It is non-trivial to determine what weight is big enough to consider both communities together. Therefore, in this work we choose to be more conservative, considering always such communities together. In fact, this is also a known problem in graph theory known as \textit{connected component} detection \cite{john1995first}. In graph theory, a connected component of a graph is a connected subgraph that is not part of any larger connected subgraph. Such components separate the vertices into disjointed sets. In the case of Figure \ref{fig:ex2}, there is one connected component, i.e., the whole graph. An advantage of this choice is that connected components can be determined in linear time $\mathcal{O}(n)$ \cite{john1995first} where \textit{n} is the number of vertices, while the Louvain algorithm runs in $\mathcal{O}(m)$ where \textit{m} is the number of edges, which is typically considered slower as the number of edges is higher than the number of vertices \cite{traag2015faster}. In this work, we filter out communities with one aircraft. 

Following Equation 11, we can determine the contribution of the community to the overall sector complexity, which we define as the sum of contributions for the individual aircraft in the community. Determining when a community is in fact a complex community is not trivial. Complexity has often been linked to the workload of controllers \cite{chatterji2001measures}, which can be subjective \cite{marchitto2016air, ayaz2010cognitive,weiland2013real}. In this work, we do not attempt to make any claims to relate the complexity indicators with the workload of controllers, we merely present a methodology that provides granular information of complexity given the definition of the aforementioned complexity indicators (which are objective). Therefore, in this work, we propose setting a contribution threshold above which a community is deemed to be a complex community. In such a way we maintain some flexibility in defining complex communities to better fit decision makers at the present structure of ATM, allowing each user to set a particular threshold. Formally, for a community $\mathcal{C}$ in time step $t$, we have:
\begin{equation}
    complex(\mathcal{C},t) = \begin{cases}
True &\text{if $\sum_{i \in \mathcal{C}} c(i,t) \geq thresh$} \\
False &\text{else}
\end{cases}
\end{equation}
where $thresh$ is the user/problem specific threshold.

So far, we have defined complex communities only for one time step $t$. However, as we are looking at a time window (here we assume that the time window is a series of discrete time steps), the continuous evolution of traffic complexity should be analyzed. We have determined three generic events that could happen: appearance, disappearance and evolution. 

Appearance and disappearance of a complex community are defined as the time steps in which the community started being and stopped being a complex community. These two events can be trivially determined by applying Equation 12 for the length of the time window. On the other hand, the evolution of complex communities requires more consideration. During the period of time when a complex community exists, new aircraft might join it, i.e., at least one existing aircraft of the community forms an interdependency with an aircraft outside of it; or leave it, i.e., an existing aircraft stops having any interdependencies with the other aircraft of the community. In such cases, the community should be considered the same, which in this work we will refer to as having the same \textit{label}. Therefore, the problem of community evolution can be seen as determining how community labels are maintained in time. 

In order to formalize the evolution of complex communities we propose an algorithm based on the Jaccard similarity. This method, formally defined below for two arbitrary sets A and B:
\begin{equation}
    \mathcal{J}(A,B) = \frac{|A \cap B|}{|A \cup B|}
\end{equation}
measures similarity of sets as the size of intersection between the size of the union of these sets. The range of $\mathcal{J}$ is between 0 and 1, with 0 meaning that the intersection is empty, i.e., the sets have no common elements. In this work, we utilize Jaccard similarity to determine the evolution of complex communities through time. For an arbitrary community $\mathcal{C}$ in time $t$, we determine if there are communities in $t-1$ that are similar to it. If in fact there are multiple communities that have a non-zero similarity to $\mathcal{C}$, then the one with the biggest similarity score is determined to have the same label as community $\mathcal{C}$. In more concise terms, if communities share some members in consecutive time steps, they are defined to share the same label feature of this algorithm is that it is only necessary to look in the previous time step, as labels can be propagated through time. Furthermore, it is trivial to determine at what time step aircraft joined or left an existing community.

If there are no similar communities in $t-1$ then community $\mathcal{C}$ is a new label in the set of complex communities for time window we are studying. Consequently, we can define all three generic events for complex communities in terms of labels: appearance is the first time when a label is present and disappearance is the last time when the label is present. The whole algorithm is described in Algorithm \ref{alg:alg1}.

\begin{algorithm}
\caption{Detection of Complex Communities}\label{alg:alg1}
\begin{algorithmic}
\State $t \gets [1,...,T]$ \Comment{{\fontsize{5}{4}\selectfont Time window that we are studying}}
\State $Comm_{complex} \gets \{\}$ \Comment{{\fontsize{5}{4}\selectfont Set of complex communities, initialized as empty}}
\While{$t \leq T$}
\Require $G_t$ \Comment{{\fontsize{5}{4}\selectfont Assume we have the graph induced by the traffic in t}}
\State $Comm_t \gets communities(G_t)$ \Comment{{\fontsize{5}{4}\selectfont Find all communities for the current time step}}
\ForEach {$C \in Comm_t$}
\If {$complex(C,t)$} \Comment{{\fontsize{5}{4}\selectfont Equation 12}}
\Require $Comm_{complex}[t-1]$\Comment{{\fontsize{5}{4}\selectfont Get complex communities in t-1}}
\ForEach{$C_{t-1} \in Comm_{complex}[t-1]$}
\State $similar \gets \mathcal{J}(C,C_{t-1})$ 
\EndFor
\State $similar_{max} \gets argmax(similar)$ \Comment{{\fontsize{5}{4}\selectfont Find most similar community}}
\If $\text{No } similar_{max}$ \Comment{{\fontsize{5}{4}\selectfont No similar communities found}}
\State $\text{Add C to }Comm_{complex}$ \Comment{{\fontsize{5}{4}\selectfont Add new label to complex communities}}
\Else 
\State $\text{Update } similar_{max}$ \Comment{{\fontsize{5}{4}\selectfont Most similar community gets data from C with}}\\
\Comment{{\fontsize{5}{4}\selectfont added, removed members and time step when this happened}}
\State $\text{Update } Comm_{complex}$
\EndIf
\EndIf
\EndFor 
\EndWhile
\State \Return $Comm_{complex}$
\end{algorithmic}
\end{algorithm}
\begin{figure}[h]
\centering
\includegraphics[width=1.0\linewidth]{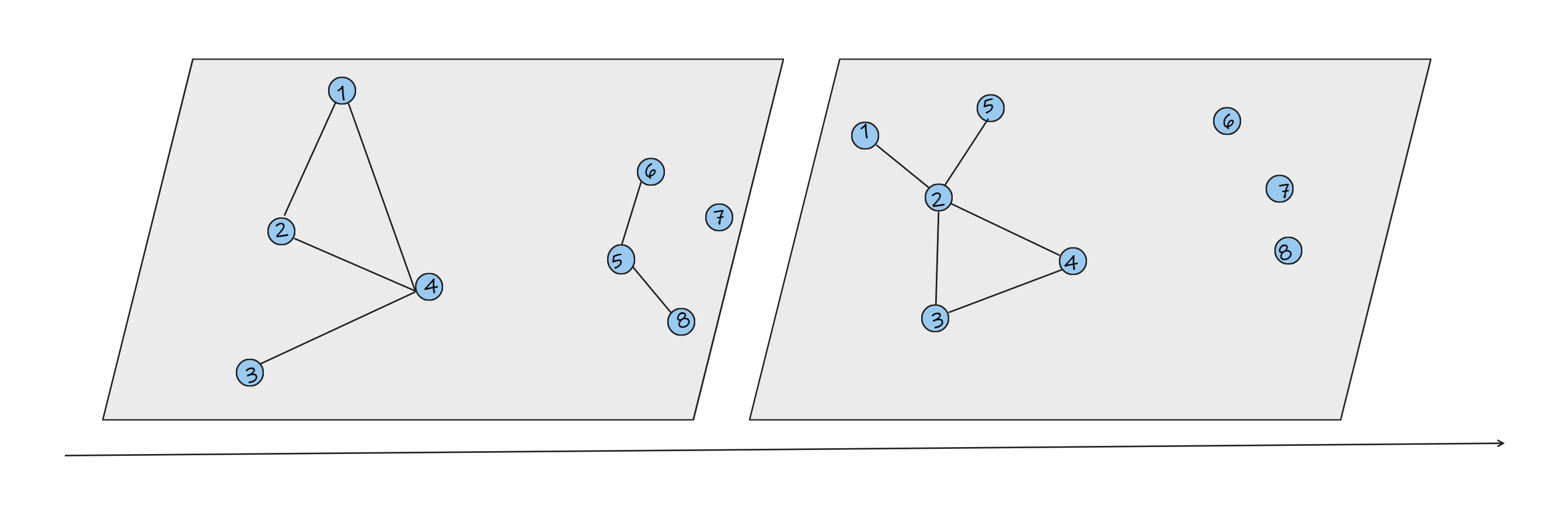}
\caption{Illustration of complex community detection algorithm.}
\label{fig:ex3}
\end{figure}
Let us illustrate this algorithm through the example in Figure \ref{fig:ex3}. There, an arbitrary sector in two time steps is shown. In $t_1$, there are three communities, namely the community labelled $\mathcal{C}_1$ with aircraft 1-4, then community $\mathcal{C}_2$ with 5-7 and the last one is community $\mathcal{C}_3$ with a single aircraft, 8, which is filtered out. Let us assume that in $t_1$ community $\mathcal{C}_1$ is complex while the other are not. As we are at the initial time step, there are no previous time steps, thus the set of complex communities would have $\mathcal{C}_1$ with members 1-4 all added at $t_1$. 

In $t_2$, the communities have changed in terms of membership. Community $\mathcal{C}_4$ with members 1-5 is a complex community. As per the algorithm, the previous time step would be queried to find any other existing complex communities, where $\mathcal{C}_1$ would be found. Then, the Jaccard similarity would be calculated with $\mathcal{J}(\mathcal{C}_1,\mathcal{C}_4) = 0.8$. Therefore, there exists one complex community in the previous time step that has a non-zero similarity score with $\mathcal{C}_4$. This means that $\mathcal{C}_4$ received $\mathcal{C}_1$ as the label and the original community is updated to contain the new information. Thus, $\mathcal{C}_1$ now contains aircraft 1,2,3,4,5 with the former 4 aircraft being added in $t_1$ and aircraft 5 being added in $t_2$. For the sake of completeness, let us investigate what happens to community $\mathcal{C}_2$ in $t_2$. As it can be observed, the remaining aircraft that comprised $\mathcal{C}_2$ have no interdependencies with any other aircraft. Therefore, there are no communities with non-zero similarity to $\mathcal{C}_2$. In this case, if $\mathcal{C}_2$ was indeed a complex community we would be able to say that it appeared in $t_1$ with members 5,6,7 and disappeared in $t_2$.
\section{Experimental Setup}
In order to effectively showcase the algorithm, a tool was built to visualize the results. This tool was developed as a web application in Python using Dash \footnote{\url{https://plotly.com/dash/}} as a frontend. There are four main plots that are the outputs of the tool:
\begin{itemize}
    \item \textit{Complexity animation} For every time step in the window of interest, the positions, interdependencies and complexity contributions for each aircraft are shown. This information is shown as an animation through the time window. The goal of this output is to clearly convey the evolution of complexity during a particular time window. In order to visually indicate when a community is complex, the interdependencies between aircraft of this community are colored in red. 
    \item \textit{Strength indicator animation} This plot provides similar information as the previous one. However, in this only the strength indicator is shown. More specifically, for each aircraft we provide the value of the maximal weight of the pairwise interdependencies that it is part of. As the strength indicator is defined through pairwise distances, it is directly linked to conflicts and losses of separation. Thus, the goal of this plot is to show specific safety related information.
    \item \textit{Heatmap of complex communities} This output shows as a heatmap the contribution of every complex community that has existed through the duration of the time window. As the x-axis is time, the coexistence or any other time relation between complex communities can also be inferred. We also keep track of the aircraft in the sector that do not belong to a complex community, which we refer to as "Pool". All aircraft that are responsible for some of the complexity in the sector are shown there. These aircraft are also part of communities that are responsible for less than the complexity threshold. When no complex communities exist, the Pool is responsible for 100 \% of complexity.
    \item \textit{Summary table} This table shows a detailed summary of every complex community that has existed in the time window. We show relevant information such as start and end time, all members that have been part of the community and when each member was added and removed.
\end{itemize}
Lastly, there is also the possibility to generate and download a summary file for the current log file. This summary file contains the values of the input parameters, as well as statistical information regarding the number, size, duration and percentage of communities. The statistical information comprises of the mean, standard deviation and minimum and maximum values. However, such a functionality could be easily adapted or extended with respect to the needs of the practitioners.

\begin{figure}[h]
\centering
\includegraphics[width=1.0\linewidth]{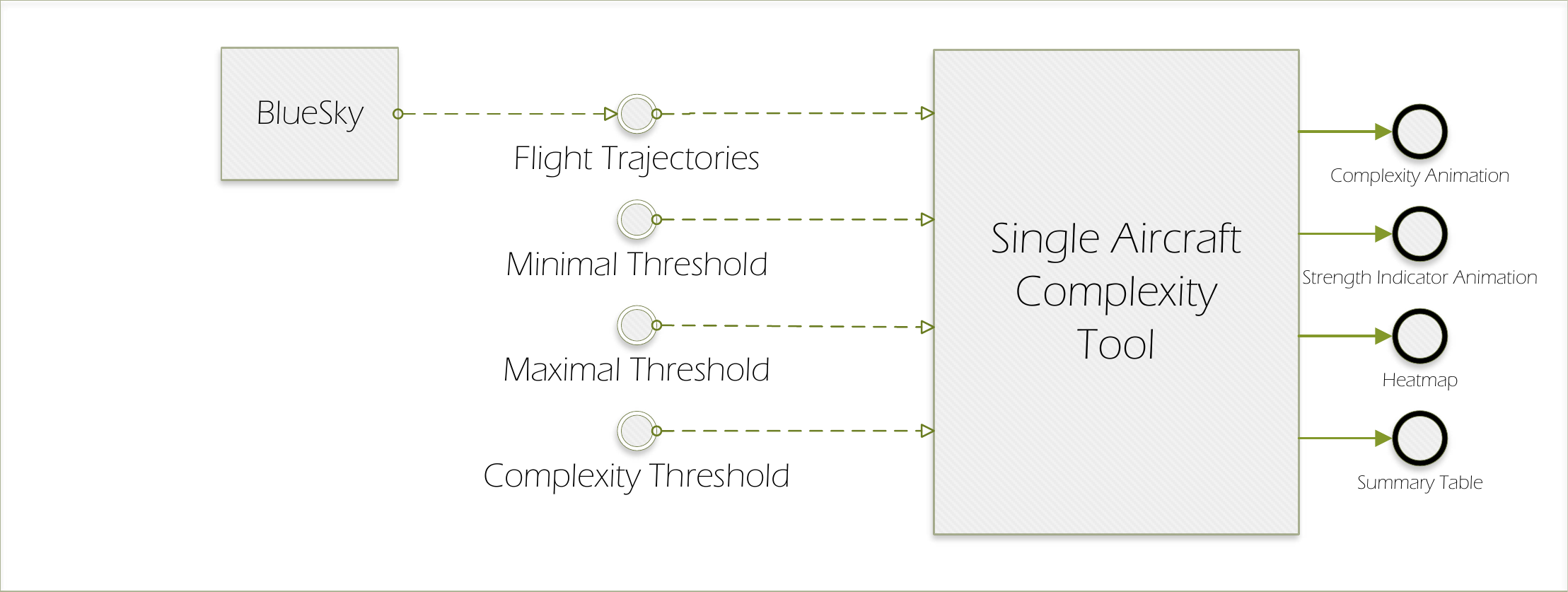}
\caption{Workflow for visualization tool.}
\label{fig:workflow}
\end{figure}

The workflow to use the tool is shown in Figure \ref{fig:workflow}. In this work, we use BlueSky \cite{hoekstra2016bluesky} as the simulation platform for the trajectories. From BlueSky the tool requires as input a file that for every time step logs the positions of every aircraft. We note that it is not a requirement to use BlueSky and the tool is not dependent on it. The file with the logged information is, however, a requirement. Furthermore, there are three more inputs to the tool, namely the minimal and maximal thresholds in order to form the interdependencies and the complexity threshold for a community to be considered complex. As this tool has been written in Python, it will be straightforward to extend the functionalities of the tool and also integrate it with other existing tools, services and infrastructure. 

\section{Results}
\subsection{Synthetic Traffic}
In this section, we will describe several scenarios based on synthetic traffic to showcase how the information provided by the methodology and the tool can be utilized. First of all, we will show a scenario where we analyze ATC decisions. In another scenario, we illustrate how ATFM decisions can affect KPAs.

\subsubsection{Pairwise Conflicts}
In this scenario, shown in Figure \ref{fig:sc1}, we start with an arbitrary sector and 5 present aircraft and the simulation lasts for 15 minutes. The trajectories have been generated in such a way, that there will be a conflict between AC2-AC3 and AC4-AC5 at some time $t$. Furthermore, the conflict between AC4-AC5 starts slightly earlier, but both conflicts will co-exist in time.

\begin{figure}[h!]
\centering
\includegraphics[width=0.8\linewidth]{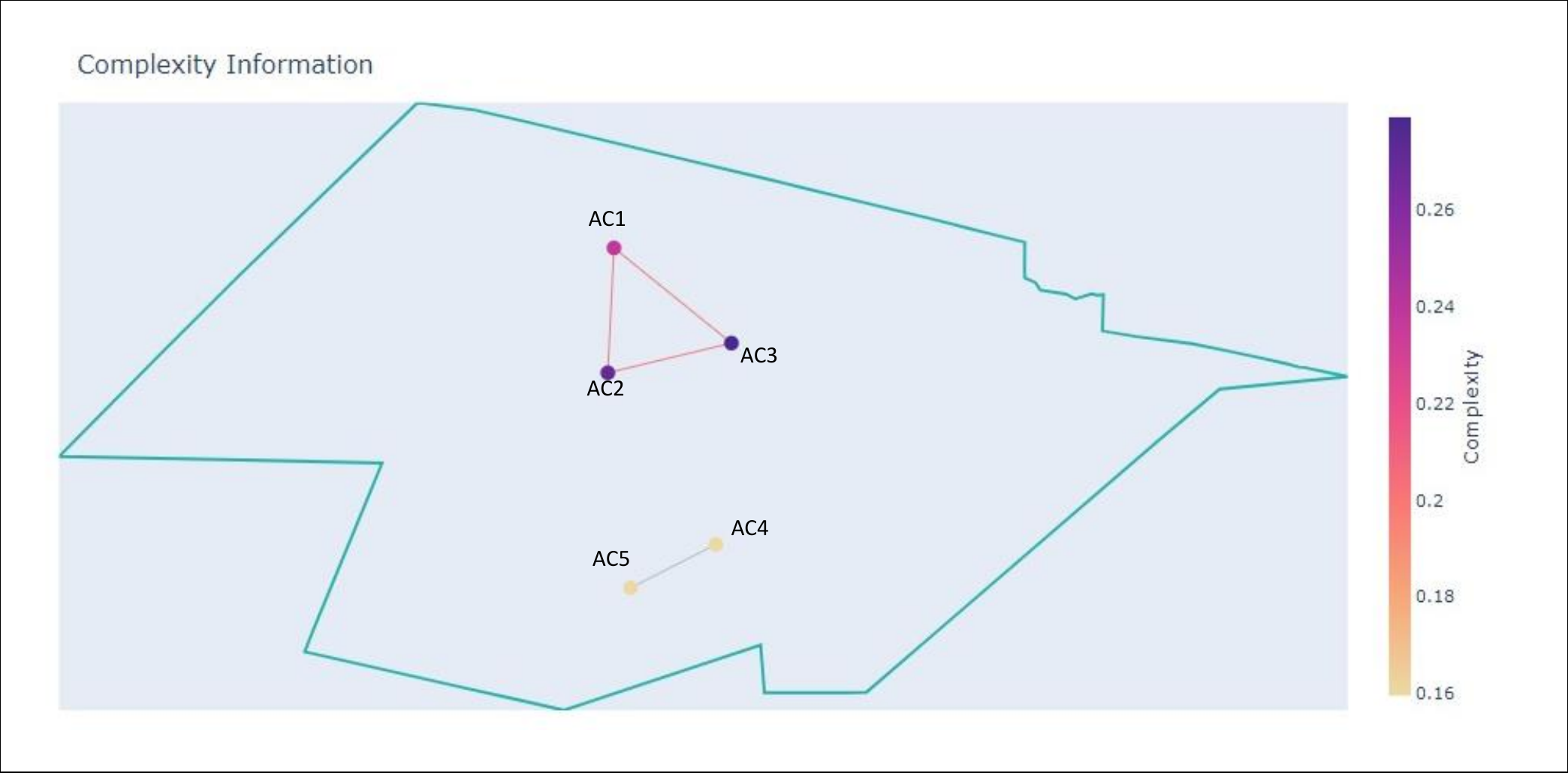}
\includegraphics[width=0.8\linewidth]{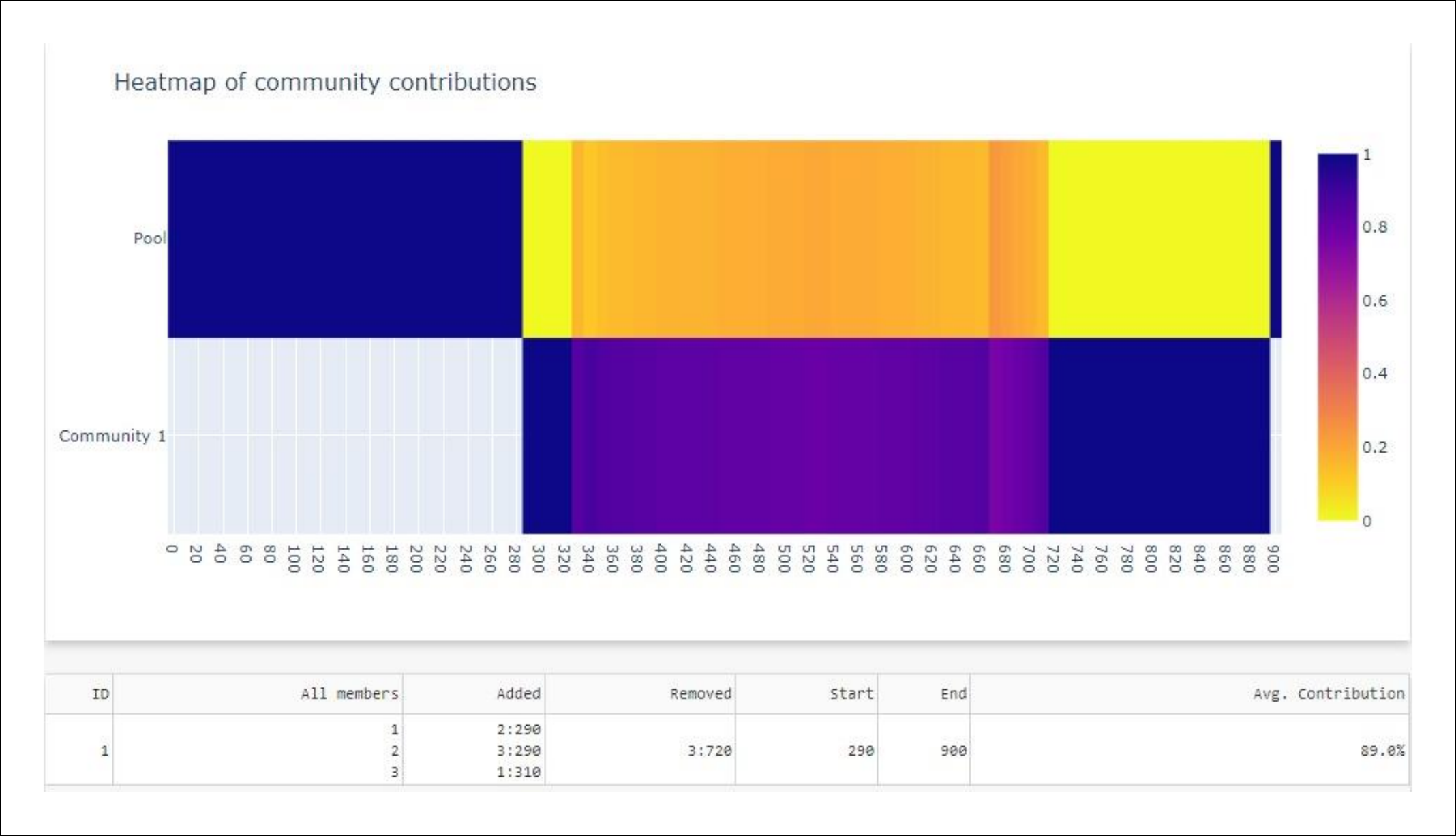}
\caption{Complexity of aircraft.}
\label{fig:sc1-comp}
\end{figure}

\begin{figure}[h]
\centering
\includegraphics[width=0.8\linewidth]{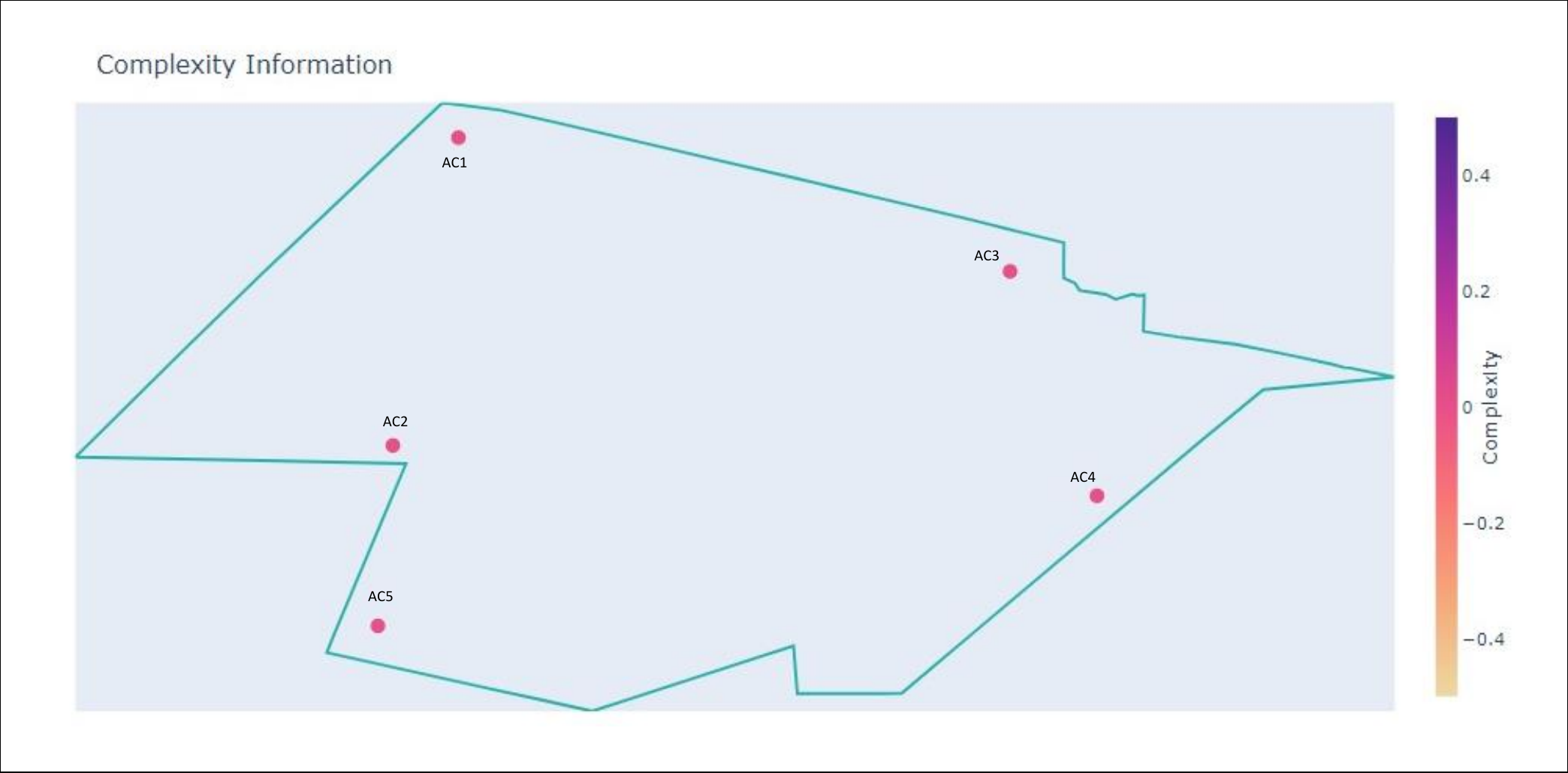}
\caption{Initial state of scenario.}
\label{fig:sc1}
\end{figure}

In such a situation, the ATCo will have to solve two conflicts that are happening around the same time and we assume that they are unable to solve them simultaneously. It is also assumed that there is enough time for ATCos to prevent losses of separation in pairwise conflicts.

\begin{figure}[h]
\centering
\includegraphics[width=0.8\linewidth]{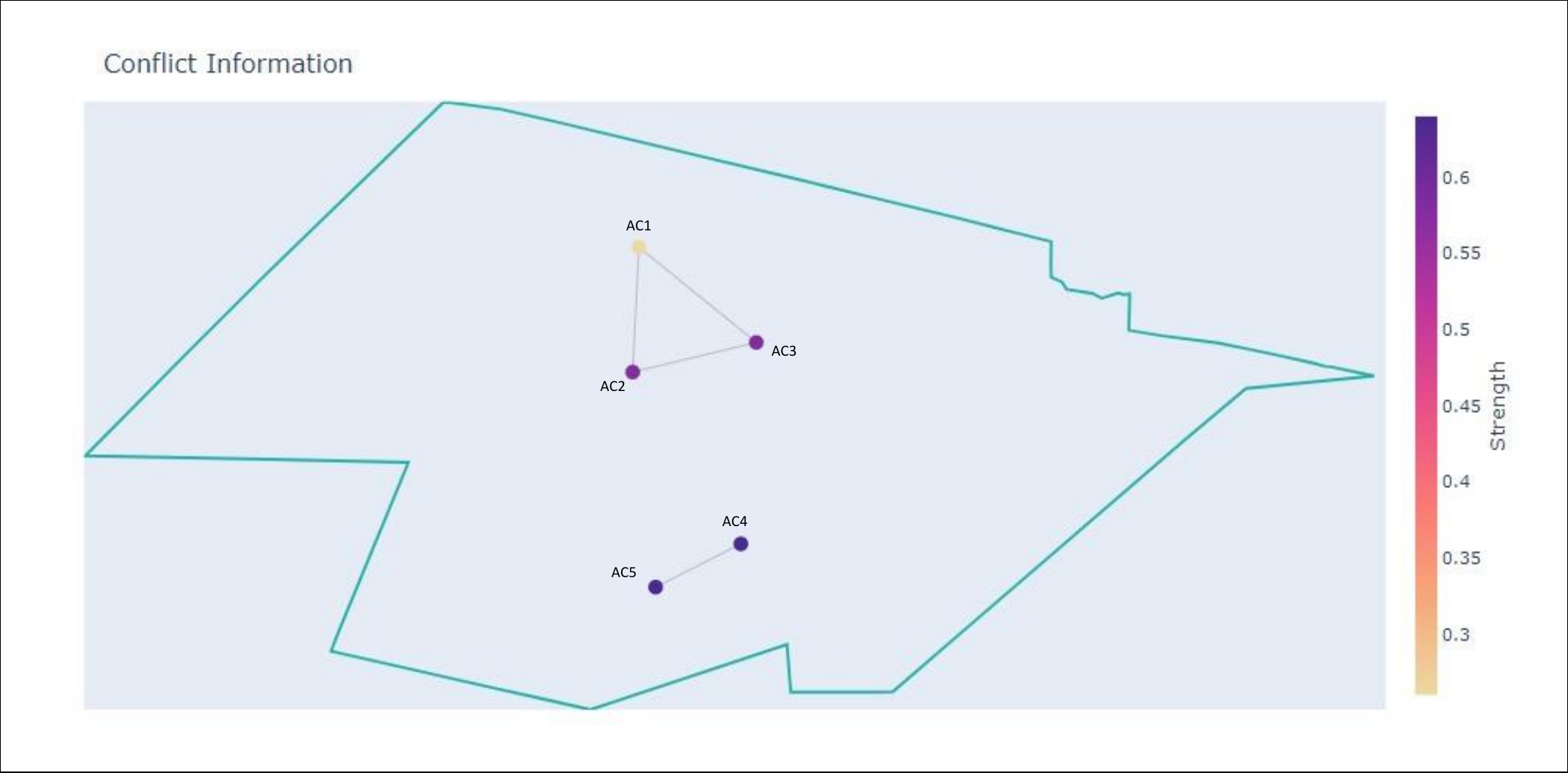}
\caption{Conflict State.}
\label{fig:sc1-conf}
\end{figure}

In Figure \ref{fig:sc1-conf} the state of the aircraft when the ATCo should have been alerted by the present Conflict Detection method is illustrated. This is shown in the tool through the Strength indicator, which directly correlates with the relative state between aircraft. The Strength of AC2, AC3 is 0.55 and AC4 and AC5 is around 0.6 (1 is a loss of separation). Typically, ATCos would solve the earlier conflict first, i.e, AC4-AC5. However, the presence of AC1 complicates this decision. Following the evolution of trajectories, AC1 will eventually create a compound conflict \cite{isufaj2022multi} with AC2 and AC3. This is not shown yet by using only the Strength indicator, however, when measuring single aircraft complexity as previously described, we can observe the following complexity situation shown in Figure \ref{fig:sc1-comp}. There, it can be seen that the community created by AC1, AC2 and AC3 is a complex community. 

The provided information should affect the decision of the ATC by considering three different KPAs. 
\begin{itemize}
    \item \textit{Safety} - First of all, if the conflict between AC4 and AC5 is solved first, it is not guaranteed that it will be done before AC1 is in conflict with AC2 and AC3. This means that the ATCos would have to solve a compound conflict. The algorithm quantifies this information and the tool presents it in such a way that clearly illustrates which aircraft form the compound conflicts and complex communities. Therefore, using the information provided by the tool, the ATCo should make the decision of to solve the pairwise conflict between AC2 and AC3 first. Furthermore, the conflicts could be resolved in such a way that at best reduces the overall complexity and at worst just avoids secondary conflicts. This information could be acquired by running the algorithm again after a resolution is proposed.
    \item \textit{Efficiency} - However, the controllers might still be able to solve the compound conflict. One way to solve it could be to force one of the aircraft to have a large deviation from its original trajectory. This solution effectively would reduce the compound conflict to a pairwise conflict. However However such a resolution would not be preferred as the aircraft that is deviated will incur delays that result in inefficient use of time and fuel. 
    \item \textit{Capacity} - Nevertheless, delays to one of the aircraft might be unavoidable. Another option could for the controller to determine that they are not able to solve these conflicts in time. Let us assume that in this case, ATC would make a request for one of these aircraft to be delayed. The ATCos will have the information that which aircraft are in conflict and which aircraft form a complex community. Consequently, delaying one of the aircraft could be done by maintaining some fairness and therefore one of the aircraft of the complex community should be delayed. Determining which of them would depend on the capabilities of the ATCos to solve two pairwise conflicts around the same time.

\end{itemize}

\subsubsection{Deconstructing Complex Communities}
\begin{figure}[h!]
\begin{minipage}[t]{0.32\textwidth}
\includegraphics[width=\linewidth,keepaspectratio=true]{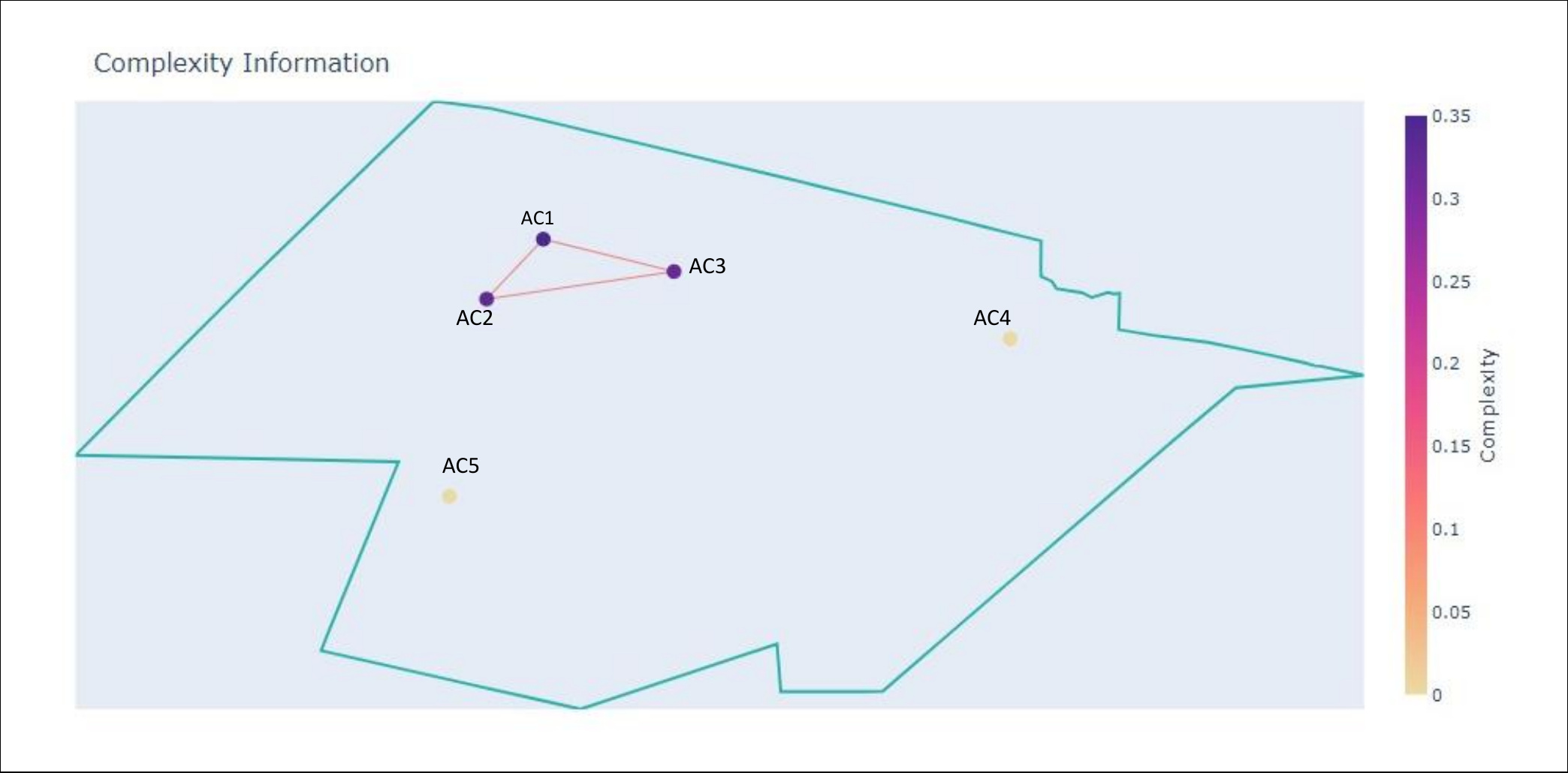}
\caption{Initial state of the sector $t_0$.}
\label{fase1}
\end{minipage}
\hspace*{\fill} % it's important not to leave blank lines before and after this command
\begin{minipage}[t]{0.32\textwidth}
\includegraphics[width=\linewidth,keepaspectratio=true]{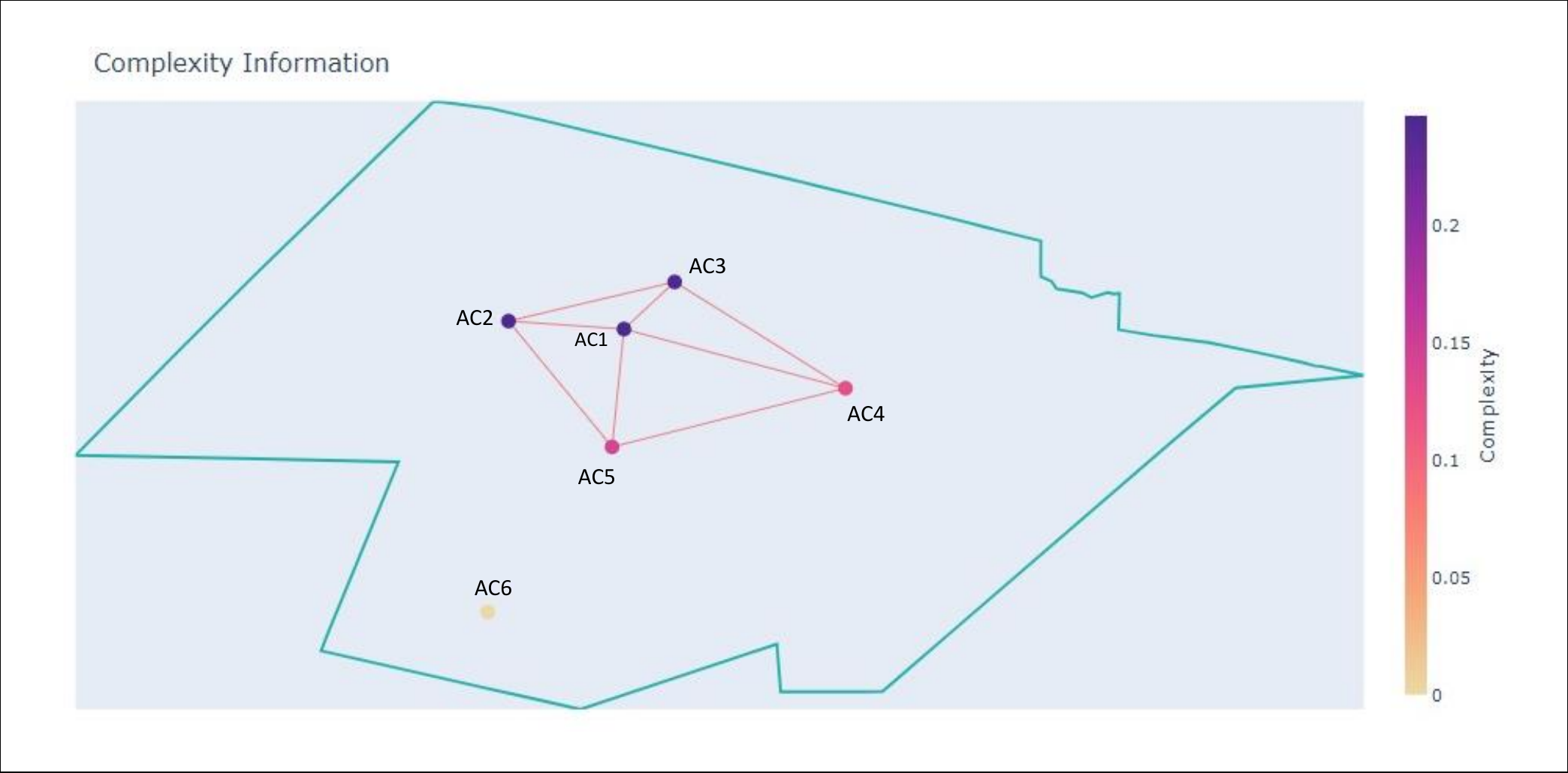}
\caption{Sector in $t_1$.}
\label{fase2}
\end{minipage}
\hspace*{\fill} % it's important not to leave blank lines before and after this command
\begin{minipage}[t]{0.32\textwidth}
\includegraphics[width=\linewidth,keepaspectratio=true]{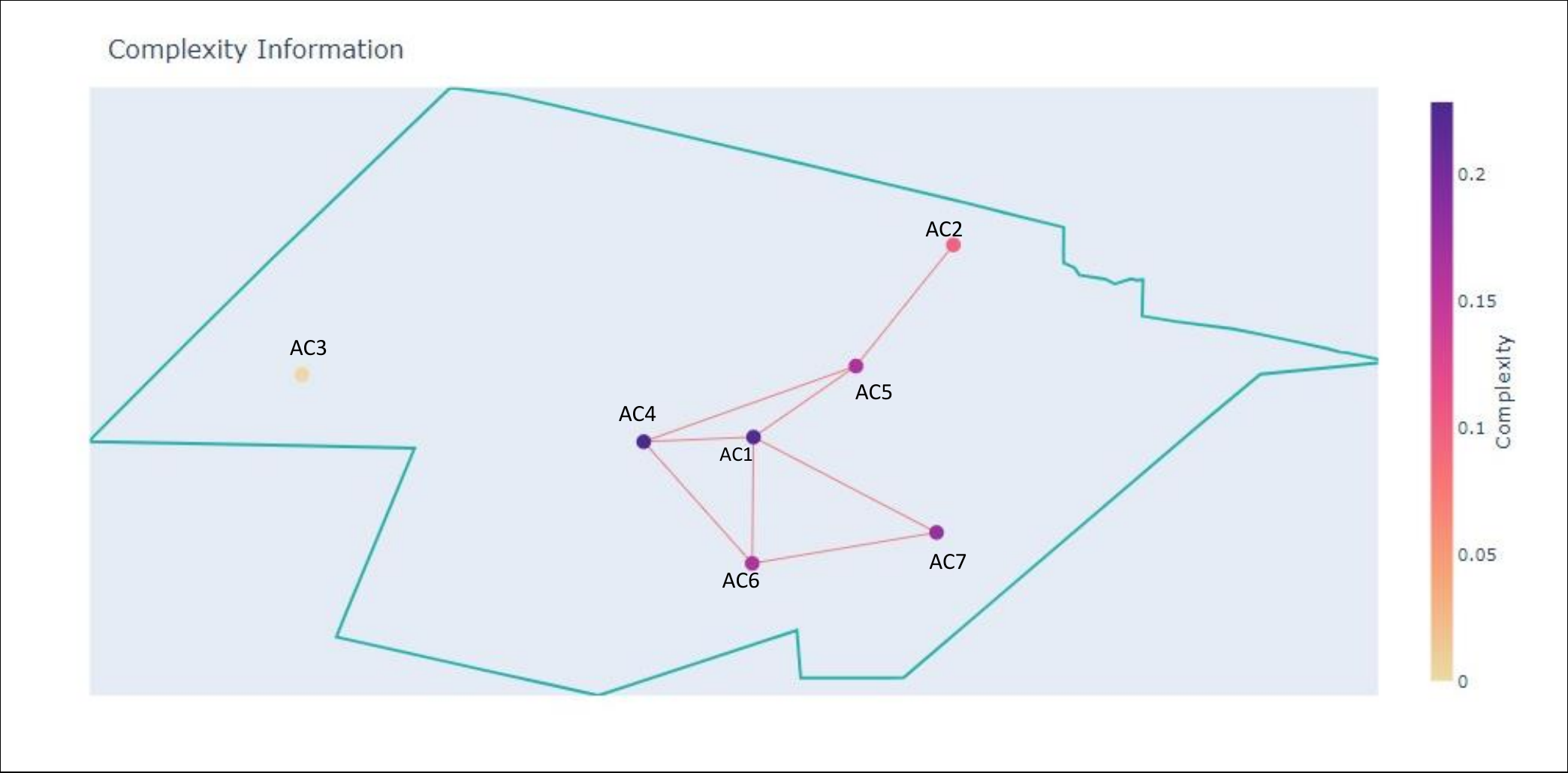}
\caption{Sector in $t_2$.}
\label{fase3}
\end{minipage}
\end{figure}

In this scenario, we are studying the same sector as in the previous section. There will be 7 aircraft present in total through the simulation which lasts 20 minutes. AC1 is present throughout the simulation and forms various interdependencies with the other aircraft. The other aircraft were generated in such a way that they intersect with the trajectory of AC1 at different time steps. There were no restrictions on whether the other aircraft can form interdependencies amongst themselves. This is illustrated in Figures \ref{fase1}, \ref{fase2} and \ref{fase3} where the sector is shown in three different time steps; initial $t_0$ where AC1 has interdependencies with AC2 and AC3 but not with AC4 and AC5. In $t_1$ where AC1 has created interdependencies with AC4 and AC5, while the remaining aircraft have also created interdendencies amongst themselves. In $t_2$ the evolution when AC1 has travelled through the sector enough to create interdependencies with AC6 and AC7. 

\begin{figure}[h]
\centering
\includegraphics[width=0.8\linewidth]{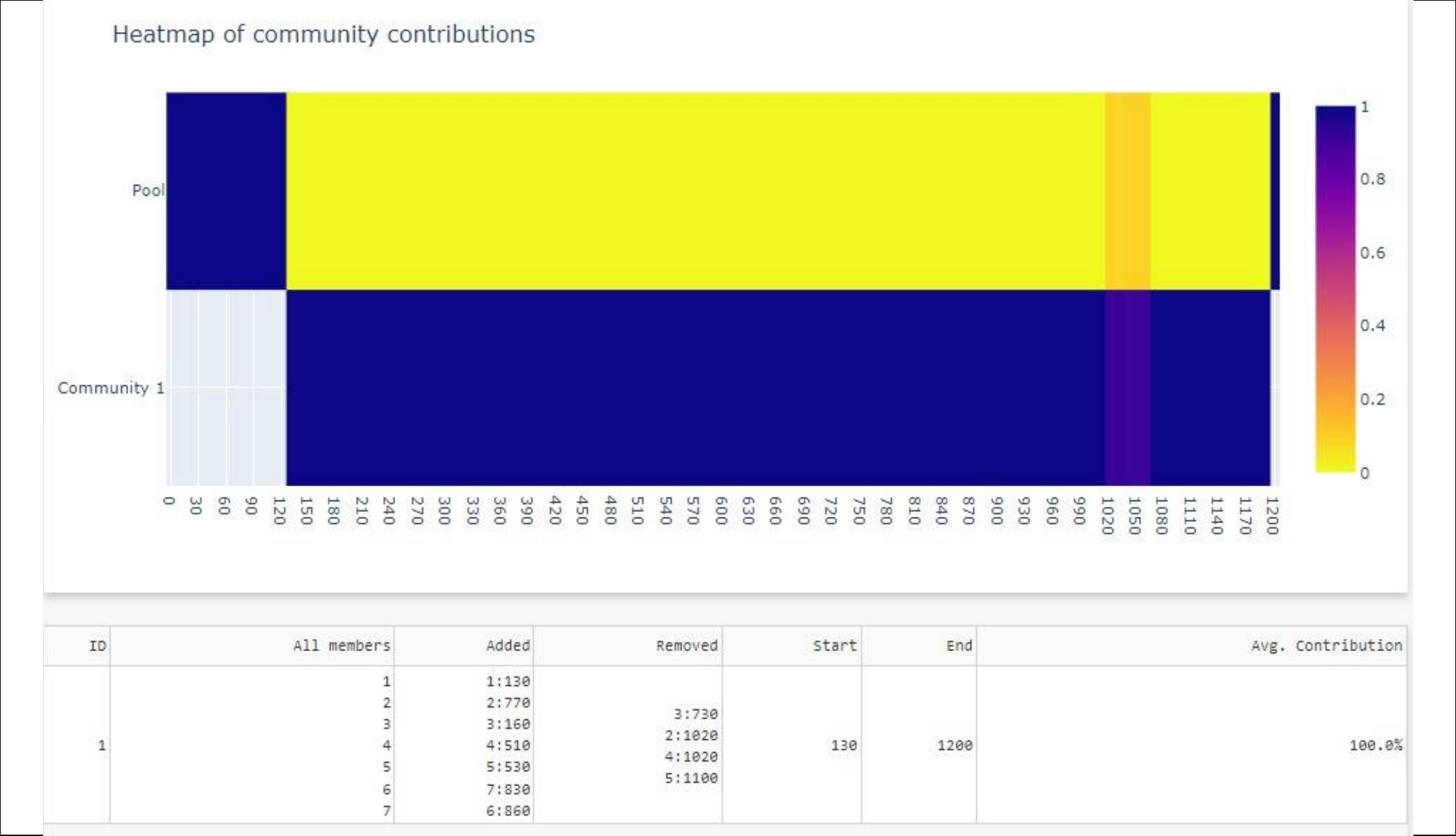}
\caption{Complexity evolution.}
\label{fig:sc2-heatmap}
\end{figure}

The complexity evolution is shown in Figure \ref{fig:sc2-heatmap}. As it can be seen, one complex community is detected that lasts from $t = 130$s to $t = 1200$s and is comprised of all the members that have passed through the sector during the simulation time. The community starts with AC1, AC2 and AC3 and during the simulation, AC1 forms interdependencies with AC4-AC7 which causes the initial community to evolve. According to Algorithm \ref{alg:alg1} communities in consequent time steps will be considered the same if they are similar enough. In this case, from Figure \ref{fase2} we observe that the community is comprised of AC1 - AC5 while in Figure \ref{fase3} the community is comprised of AC1-AC7 with the exception of AC3. Thus, AC1 is a permanent member of the complex community which causes the community to last so long. 

\begin{figure}[h!]
\begin{minipage}[t]{0.32\textwidth}
\includegraphics[width=\linewidth,keepaspectratio=true]{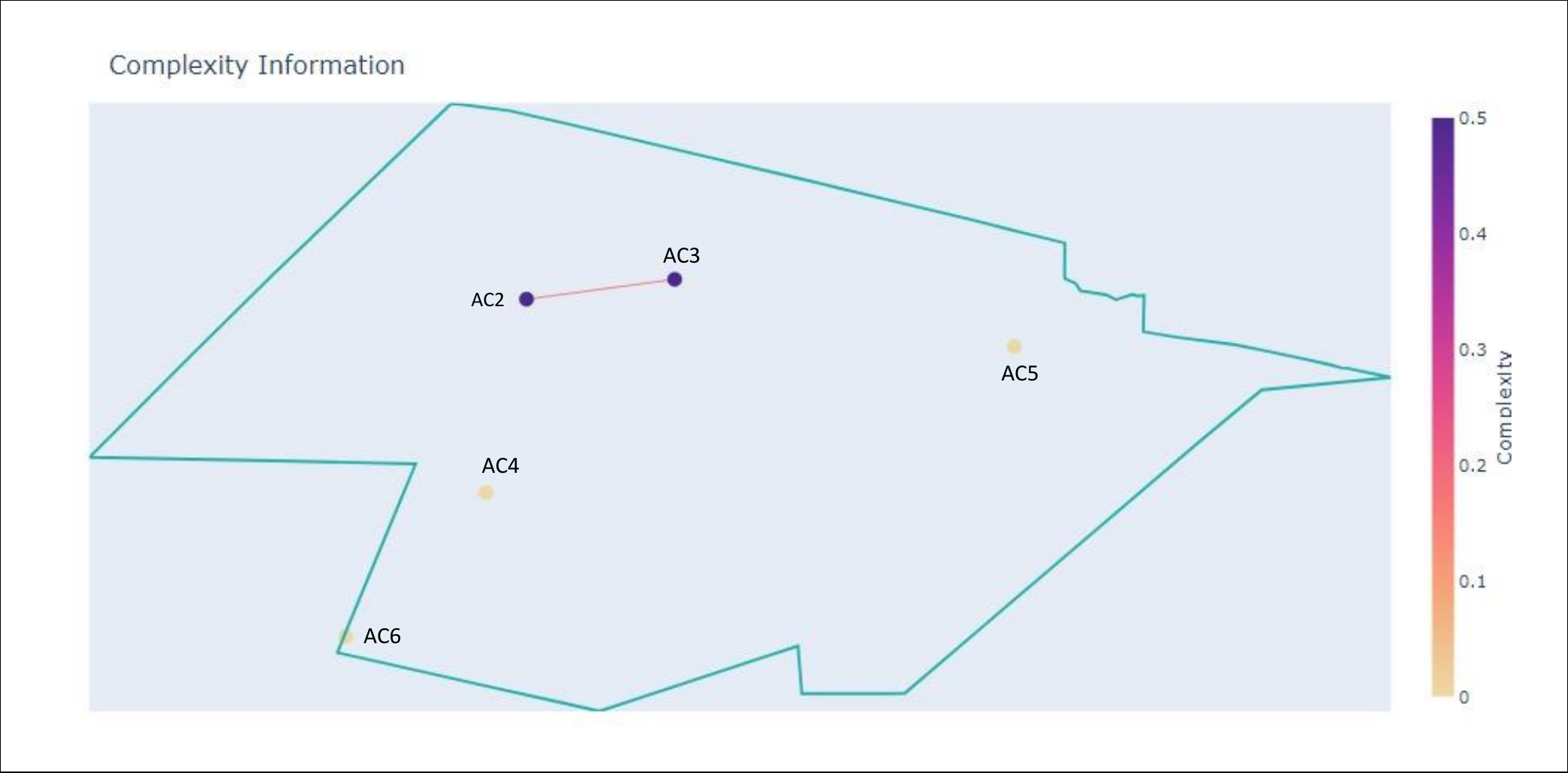}
\caption{Initial state of the sector $t_0$ without AC1.}
\label{fase11}
\end{minipage}
\hspace*{\fill} % it's important not to leave blank lines before and after this command
\begin{minipage}[t]{0.32\textwidth}
\includegraphics[width=\linewidth,keepaspectratio=true]{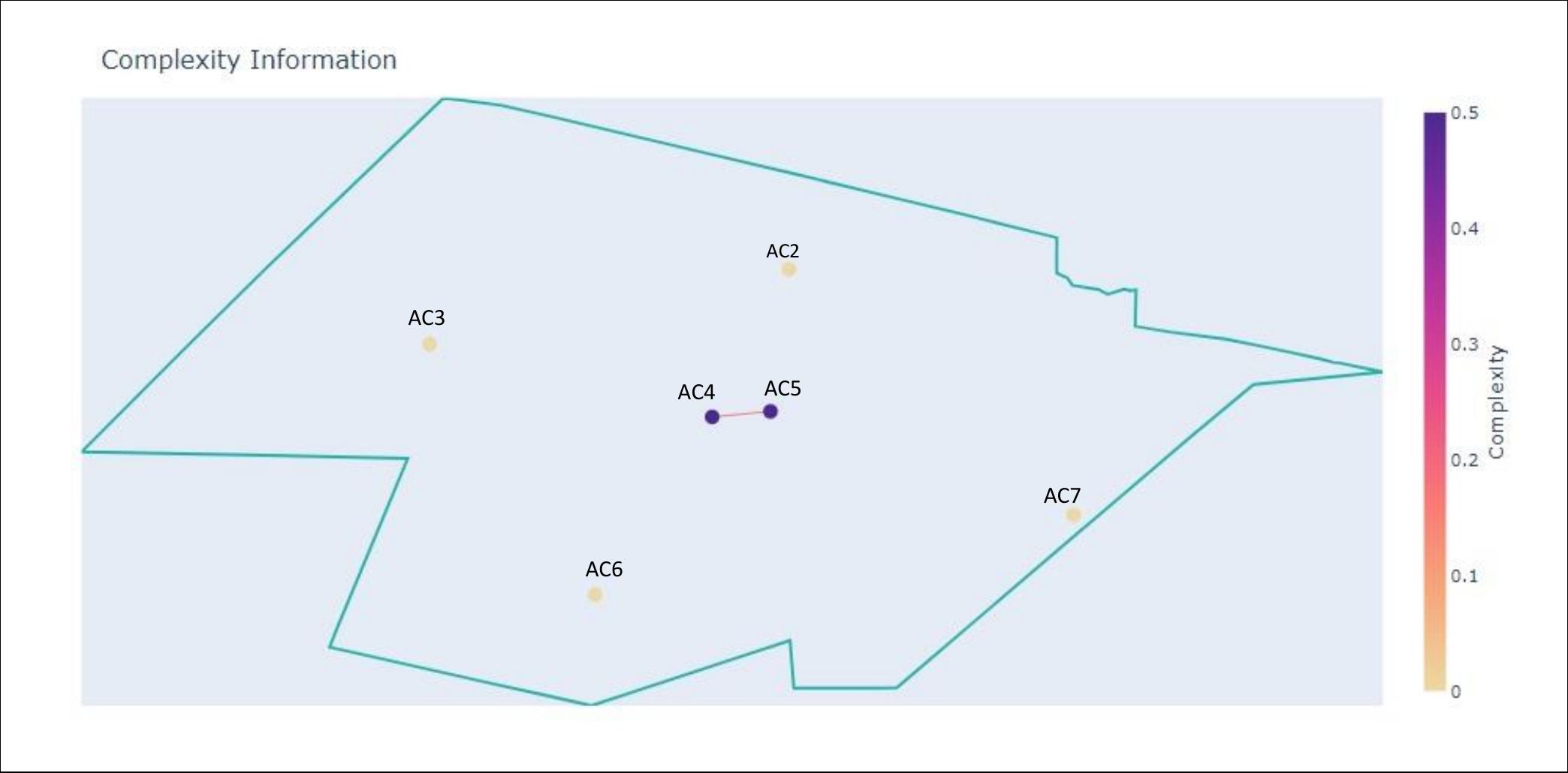}
\caption{Sector in $t_1$ without AC1.}
\label{fase22}
\end{minipage}
\hspace*{\fill} % it's important not to leave blank lines before and after this command
\begin{minipage}[t]{0.32\textwidth}
\includegraphics[width=\linewidth,keepaspectratio=true]{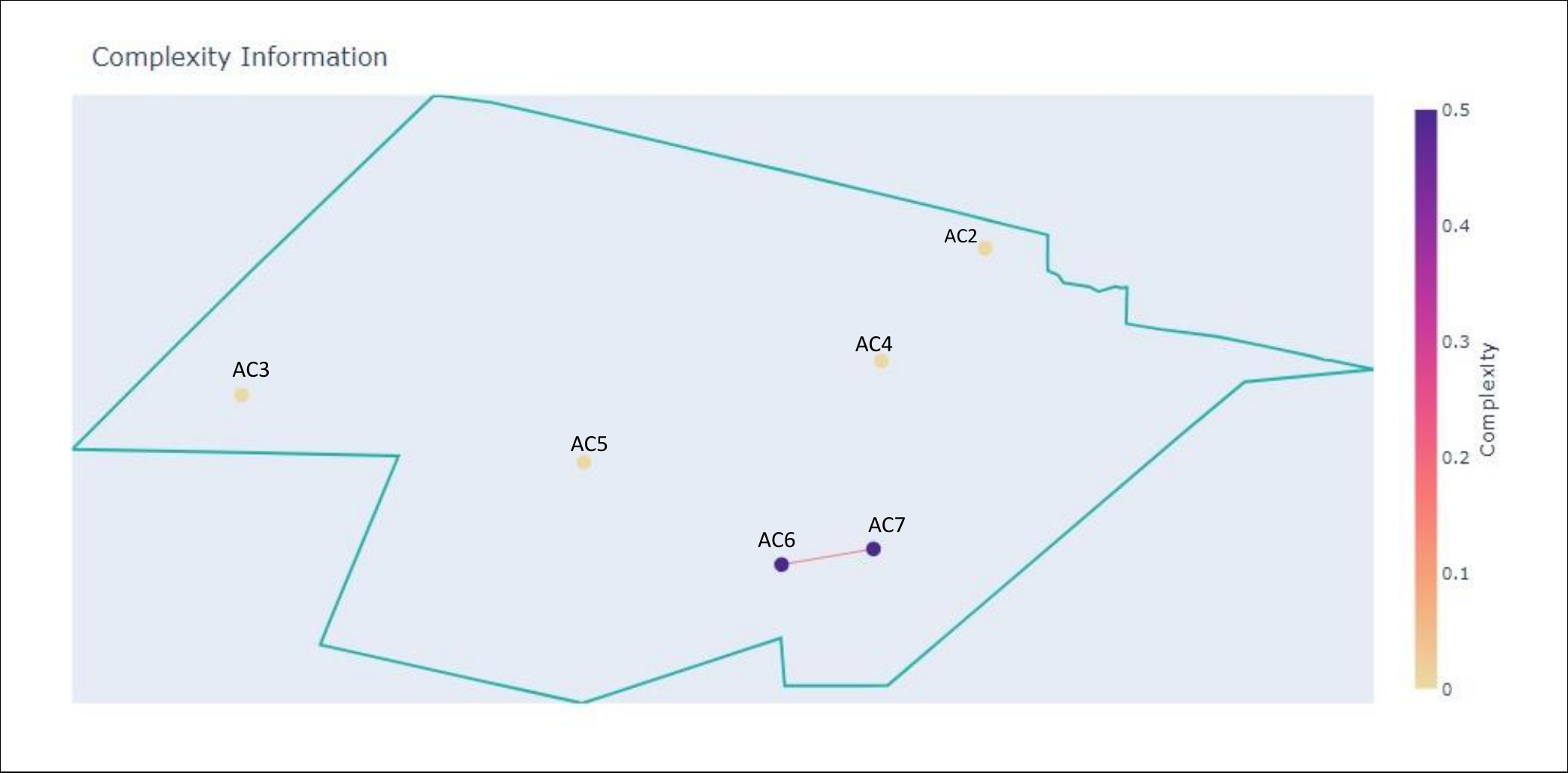}
\caption{Sector in $t_2$ without AC1.}
\label{fase33}
\end{minipage}
\end{figure}

\begin{figure}[h]
\centering
\includegraphics[width=0.8\linewidth]{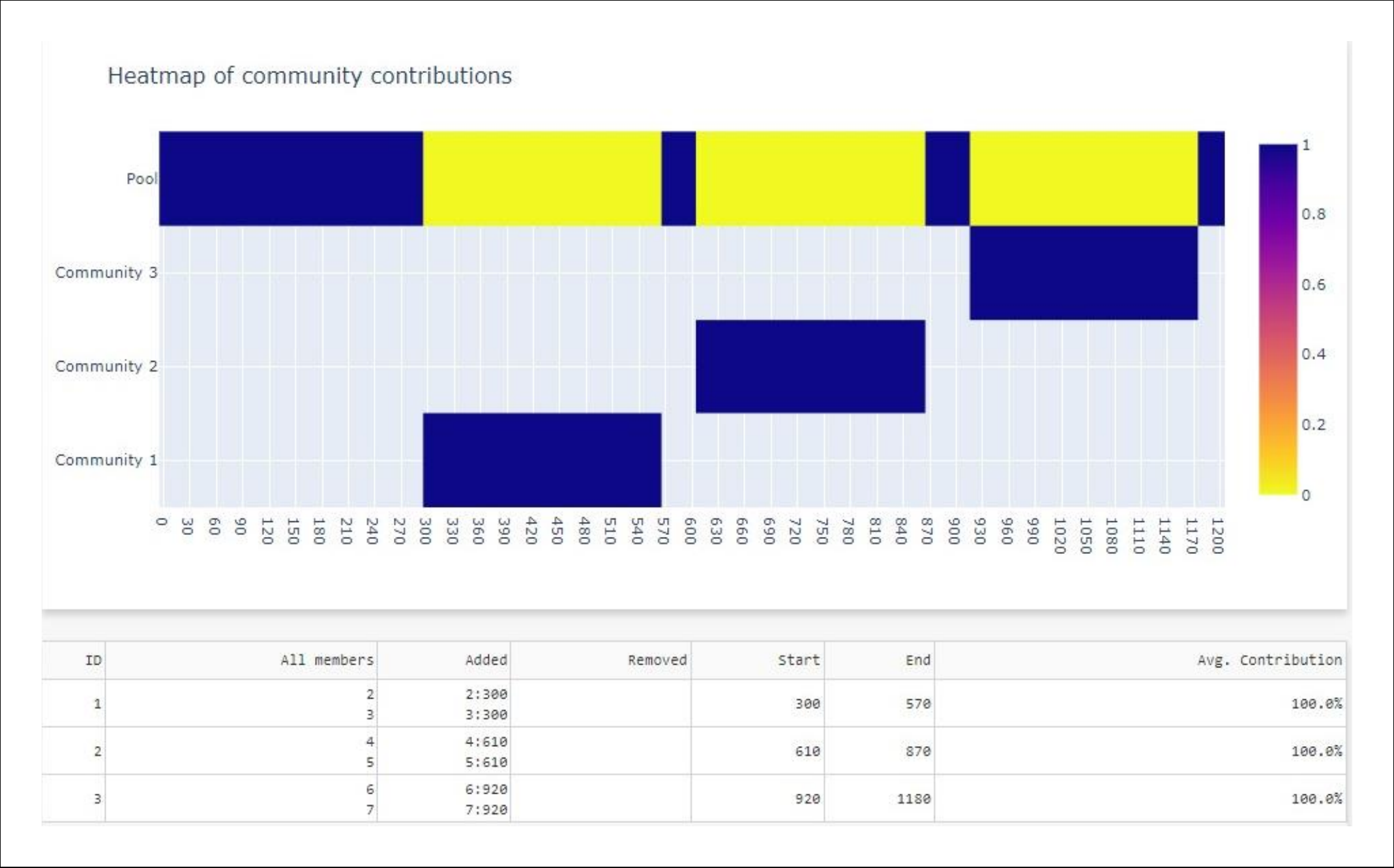}
\caption{Complexity evolution without AC1.}
\label{fig:sc2-heatmap2}
\end{figure}

The situation illustrated above could be problematic for the ATC as it demands its continous attention throughout the time window as a result of the nature of the interdependencies. Therefore, it is reasonable to expect that the ATM system should intervene to make this situation more manageable. The information that the algorithm provides through the tool could be used to better formalize and understand the consequences of the decisions. For instance, let us assume the absence of AC1 in the sector during the window of the simulation (e.g., delay, flight plan change). This is illustrated in Figures \ref{fase11}, \ref{fase22} and \ref{fase33}.
Furthermore, the complexity evolution in this case is shown in \ref{fig:sc2-heatmap2}.

In this situation the absence of AC1 has resulted in the previous complex community to be split in 3 smaller but still complex communities. It can also be noted that these communities do not overlap in time. The communities have only 2 members which causes only Strength to be a non-zero complexity indicator. This situation is less complex than the previous one where several different topologies were present in the sector. 

The lack of understanding of how and why the complex situation arises, may lead the different ATM subsystems to make arbitrary decisions which can clearly affect efficiency and fairness. Let us assume that both ATFM and ATC agree that some sort of regulation needs to be made. However, each of them could propose regulations that affect one or more aircraft. Using the information provided from the tool, the aircraft that needs to be affected from the regulations becomes evident. As previously stated, the absence of AC1 results in a more manageable situation. This is evidence that the use of the methodology proposed in this work leads to better equity and fairness at the aircraft or airline level. The objectiveness of the provided information results in a neutral tool that increases transparency and explainability of decisions made by ATM subsystems with regards to AUs. 
\subsection{Flown Trajectories}
\subsubsection{Data}
\begin{figure}[h]
\centering
\includegraphics[width=0.8\linewidth]{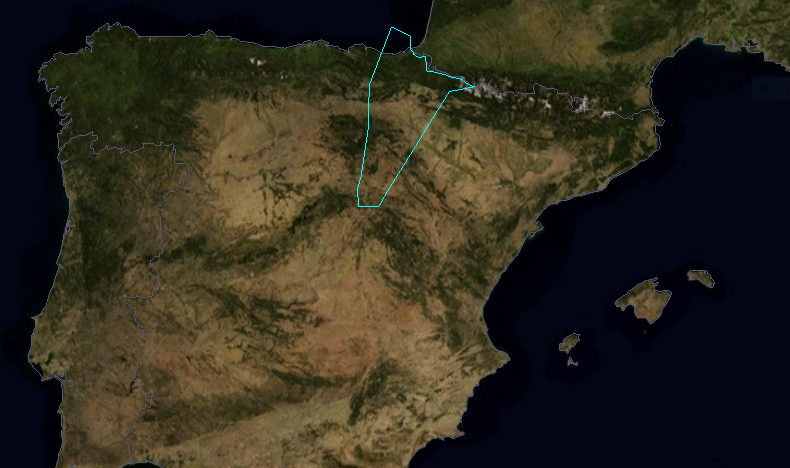}
\caption{Pamplona Upper ATC Sector.}
\label{fig:lecmpau}
\end{figure}

We will evaluate the algorithm using real historical traffic from 17.08.2019 provided by CRIDA. The available data contained flown trajectories in several sectors over Spain starting from noon and lasting for about 7 hours. In total there were 485 flights in the dataset. Furthermore, the dataset contained a list of flights that were regulated for the Pamplona Upper ATC Sector (LECMPAU), shown in Figure \ref{fig:lecmpau}. The type of regulations were time delays issued from ATC due to airspace capacity. Preliminary analysis of the data showed that ouf the 485 flights, 329 had crossed the LECMPAU sector and 82 out of those were regulated (24.9 \%). The mean delay was 14.8 minutes with a standard deviation of 11.2 minutes. The minimum delay was 1 minute while the maximum was 59 minutes.

Using the information available, we simulated two scenarios in Bluesky, one with the applied regulations and the other without the applied regulations. 

\subsection{The Effect of Regulations on Complex Communities}
In this section we will inspect how complex communities are affected by the regulations present in the data set. In order to do so, we simulate the two scenarios for the total duration of 7 hours. The trajectories with the regulations applied are the actual flown trajectories. However, we are not aware of how and when the delays were applied exactly. Thus, in order to create the trajectories without the regulations applied, we always remove the delay from the initial point of the trajectory. 

\begin{figure}[h!]
\begin{minipage}[t]{0.49\textwidth}
\includegraphics[width=1.0\linewidth,keepaspectratio=true]{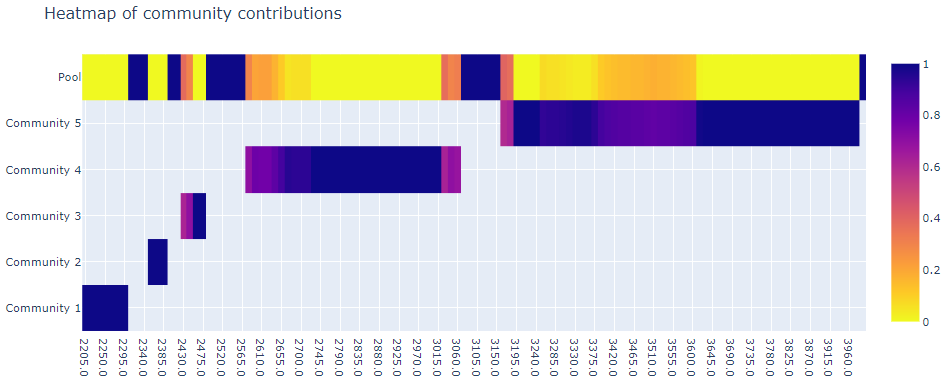}
\caption{Complex communities for the first time window with regulations.}
\label{w1d}
\end{minipage}
\hspace*{\fill} % it's important not to leave blank lines before and after this command
\begin{minipage}[t]{0.49\textwidth}
\includegraphics[width=1.0\linewidth,keepaspectratio=true]{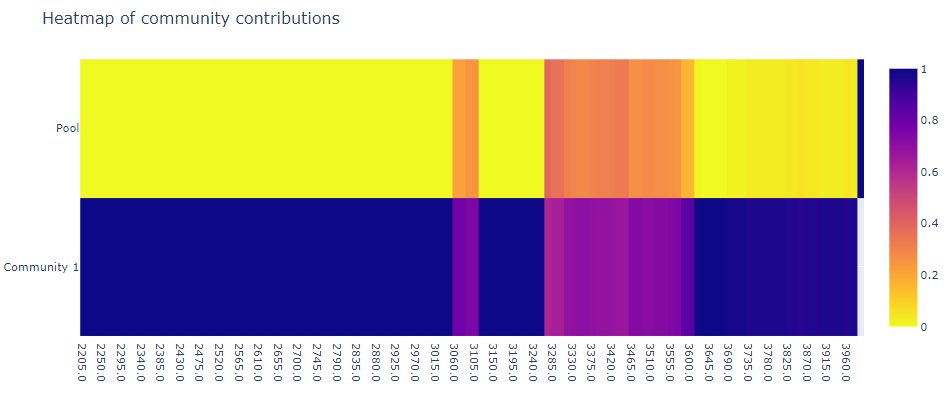}
\caption{Complex communities for the first time window without regulations.}
\label{w1nd}
\end{minipage}
\end{figure}

In order to visualize the effects, we identified three windows lasting 30 minutes each. The input parameters for the minimal and maximal distance thresholds and the complexity thresholds were set to 5 NM, 33 NM and 60\% respectively. The complex communities for the first time window are shown in Figures \ref{w1d} and \ref{w1nd}. As it can be observed, in the scenario where the regulations were not applied there exists only one community throughout the time window. Compared to the scenario where the regulations were applied, we observe 5 total complex communities. As a result of the complexity threshold being set to 60\%, these communities do not co-exist in time. This result provides evidence that the delayed aircraft were key in keeping the community in Figure \ref{w1nd} together, similar to the synthetic scenario elaborated in Section 4.1.2. The presence of only one complex community indicates that all relevant aircraft in the sector have interdependencies with them, which is also suggested by the colour of the heatmap, with the community being responsible for 100\% of the complexity for the majority of the time window. This topology is evidence of a very complex situation. Through the regulations, we can observe that the complexity is divided in time between several communities. Communities 1,2 and 3 have a relatively short duration, however they are responsible for all of the complexity in the sector. This suggests that the controller is provided elaborated information about which areas of the sector are causing the complexity present in the sector. Such information is further supported by the other outputs of the tool, e.g., the animation of single aircraft complexity contributions. This scenario serves to further illustrate two advantages of the proposed methodology: it increases the transparency of decisions made by different sub-systems of ATM and provides a framework through which to formalize the need for regulations (in this case from the point of view of ATC).

\begin{figure}[h!]
\begin{minipage}[t]{0.49\textwidth}
\includegraphics[width=1.0\linewidth,keepaspectratio=true]{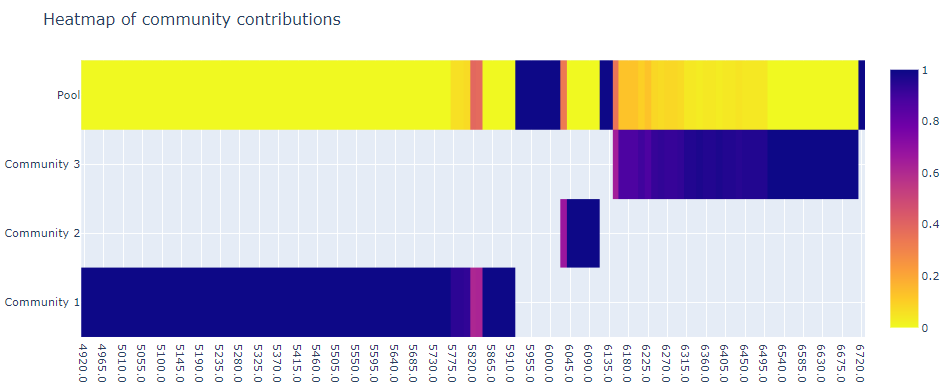}
\caption{Complex communities for the second time window with regulations.}
\label{w2d}
\end{minipage}
\hspace*{\fill} % it's important not to leave blank lines before and after this command
\begin{minipage}[t]{0.49\textwidth}
\includegraphics[width=1.0\linewidth,keepaspectratio=true]{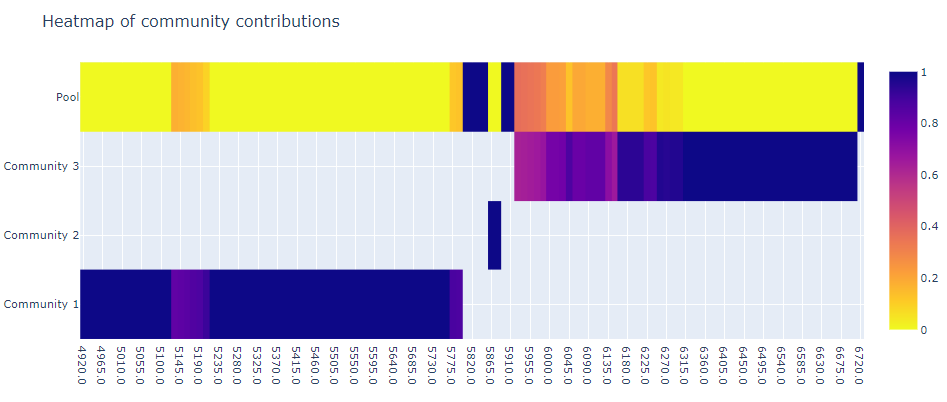}
\caption{Complex communities for the second time window without regulations.}
\label{w2nd}
\end{minipage}
\end{figure}

\begin{figure}[h!]
\begin{minipage}[t]{0.49\textwidth}
\includegraphics[width=1.0\linewidth,keepaspectratio=true]{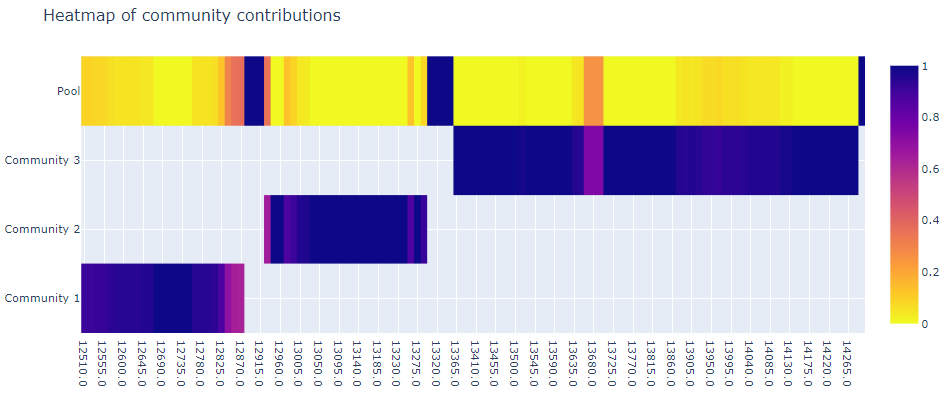}
\caption{Complex communities for the third time window with regulations.}
\label{w3d}
\end{minipage}
\hspace*{\fill} % it's important not to leave blank lines before and after this command
\begin{minipage}[t]{0.49\textwidth}
\includegraphics[width=1.0\linewidth,keepaspectratio=true]{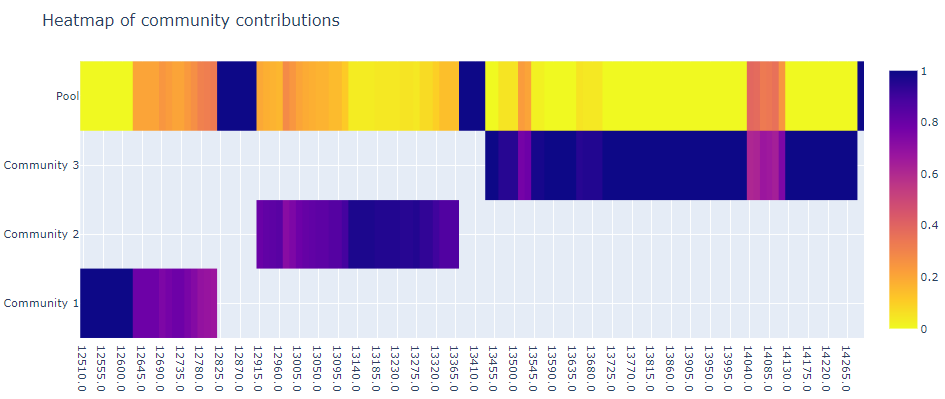}
\caption{Complex communities for the third time window without regulations.}
\label{w3nd}
\end{minipage}
\end{figure}

However, as it can be seen in the second (Figures \ref{w2d} and \ref{w2nd}) and third (Figures \ref{w3d} and \ref{w3nd}) time windows, this is not the full story. There, we can observe that the regulations change the composition of complex communities only slightly. These results suggest that complexity largely remains the same in both scenarios (i.e., with and without regulations applied). Furthermore, it is interesting to note that during these time window the difference in occupancy between the two scenarios was the highest. In the scenario without the applied regulations there were consistently 5-7 more aircraft than in the one with the regulation applied. Such a result echoes claims made in other works \cite{isufaj2021spatiotemporal, delahaye2000air} that the occupancy of sectors does not provide elaborated information about the complexity of the traffic. 

Moreover, it is worth noticing that in the scenarios without the regulations applied, the Pool tends to be responsible for more complexity. This means that the delayed aircraft were largely not present in the complex communities, but in the surrounding traffic. Such a distribution of complexity is evidence of the granularity of the information provided by the proposed algorithm, as it is able to capture subtle differences in sector complexity, which could explain in a nuanced way why the particular regulations were applied, something that sector occupancy cannot provide.

\subsubsection{Sensitivity Analysis}
The algorithm proposed in this work has 3 main parameters: minimal and maximal interdependency thresholds and the threshold for a community to be considered complex. These parameters should depend on the problem setting and also the individual user, however it is important to understand how they affect the output of the algorithm. We conduct a sensitivity analysis, which investigates how the output of a system can be attributed to its inputs. More specifically, we use the Sobol method \cite{sobol2001global,saltelli2008global} which is a variance-based global sensitivity analysis. The variance of the output is decomposed into fractions which are attributed to the inputs. The main advantage of using this method lies in the fact that it deals with nonlinear responses and it can measure the interactions between input parameters.
\begin{figure}[h!]
\centering
\includegraphics[width=1.0\linewidth]{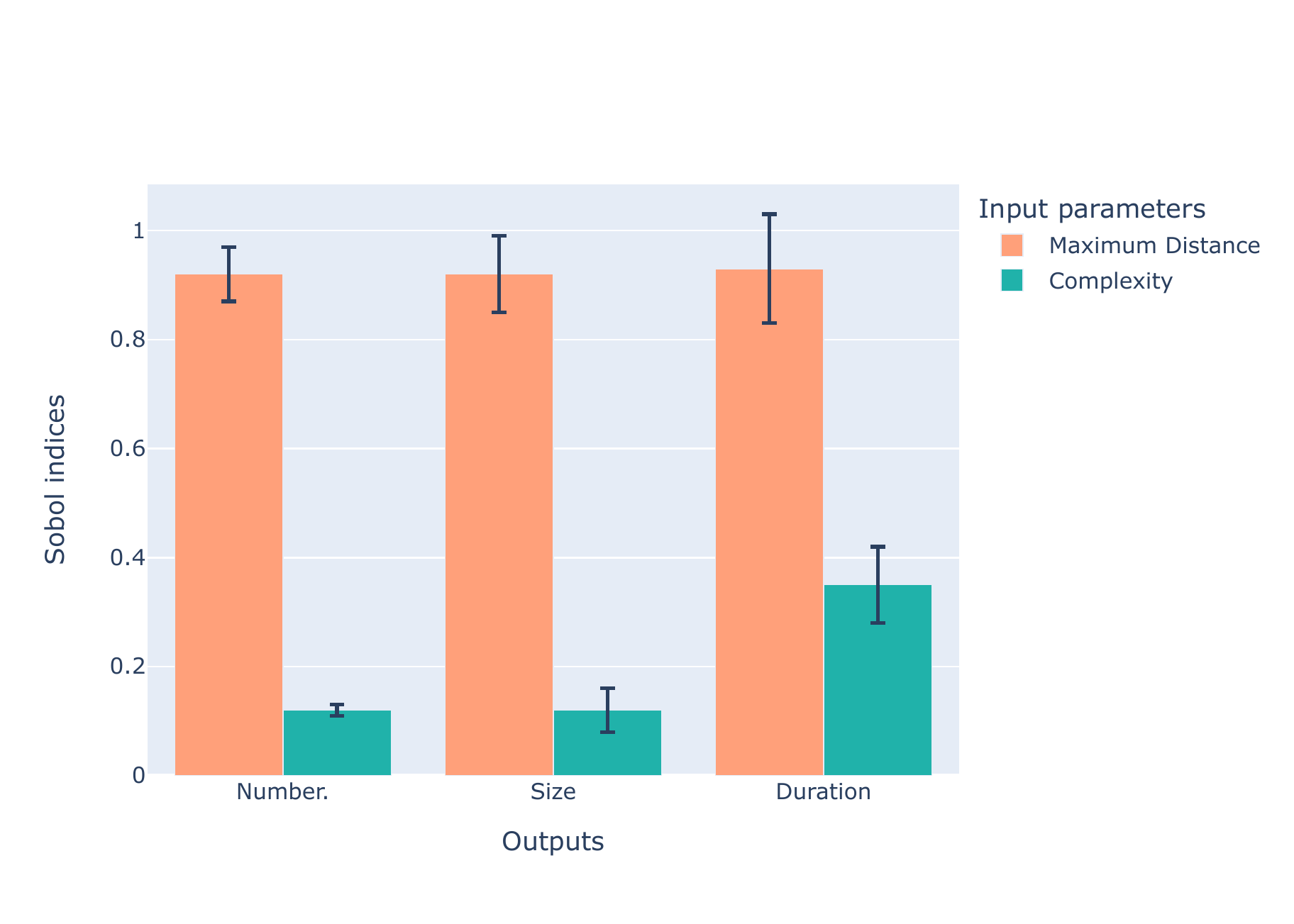}
\caption{The Total Sensitivity Analysis using the Sobol Method .}
\label{fig:sobol}
\end{figure}
In order to perform a Sobol analysis, a parameter sequence is generated, which in this work returns a Sobol sequence using Saltelli's sampling scheme \cite{saltelli2008global}. A Sobol sequence is a quasi-randomized, low-discrepancy sequence that samples the space more uniformly than a completely random sequence. The Saltelli scheme extends this sequence in a way that reduces the error rates in the calculations. To understand how the variance of the output can be attributed to the input tarameters and the interaction between each of them, the total order, first and second order sensitivity indices are calculated. The first order sensitivity indices are used to measure the fractional contribution of a single parameter to the output. Second order sensitivity anaylsis are used to measure the contribution of parameter interactions to the output variance. The total sensitivity indices take into account all the previous indices.

In this work, the sensitiviy analysis will be conducted using the regulated data described in Section 4.2.1. In order to reduce the computational burden of the analysis, motivated furthermore by the fact that this data has been of particular importance for the ATC, we keep a fixed minimal distance threshold of 5 NM. The ranges for the maximal distance threshold and the complexity threshold were $[15, 75]$ NM and $[40 \%, 100 \%]$ respectively. To perform the analysis 6200 different combinations were generated. Lastly, measured the response of three different outputs of the algorithm: number of communities, median size of communities (number of total members) and median duration of the communities.
\begin{figure}[h!]
\begin{minipage}[t]{0.49\textwidth}
\includegraphics[width=1.0\linewidth,keepaspectratio=true]{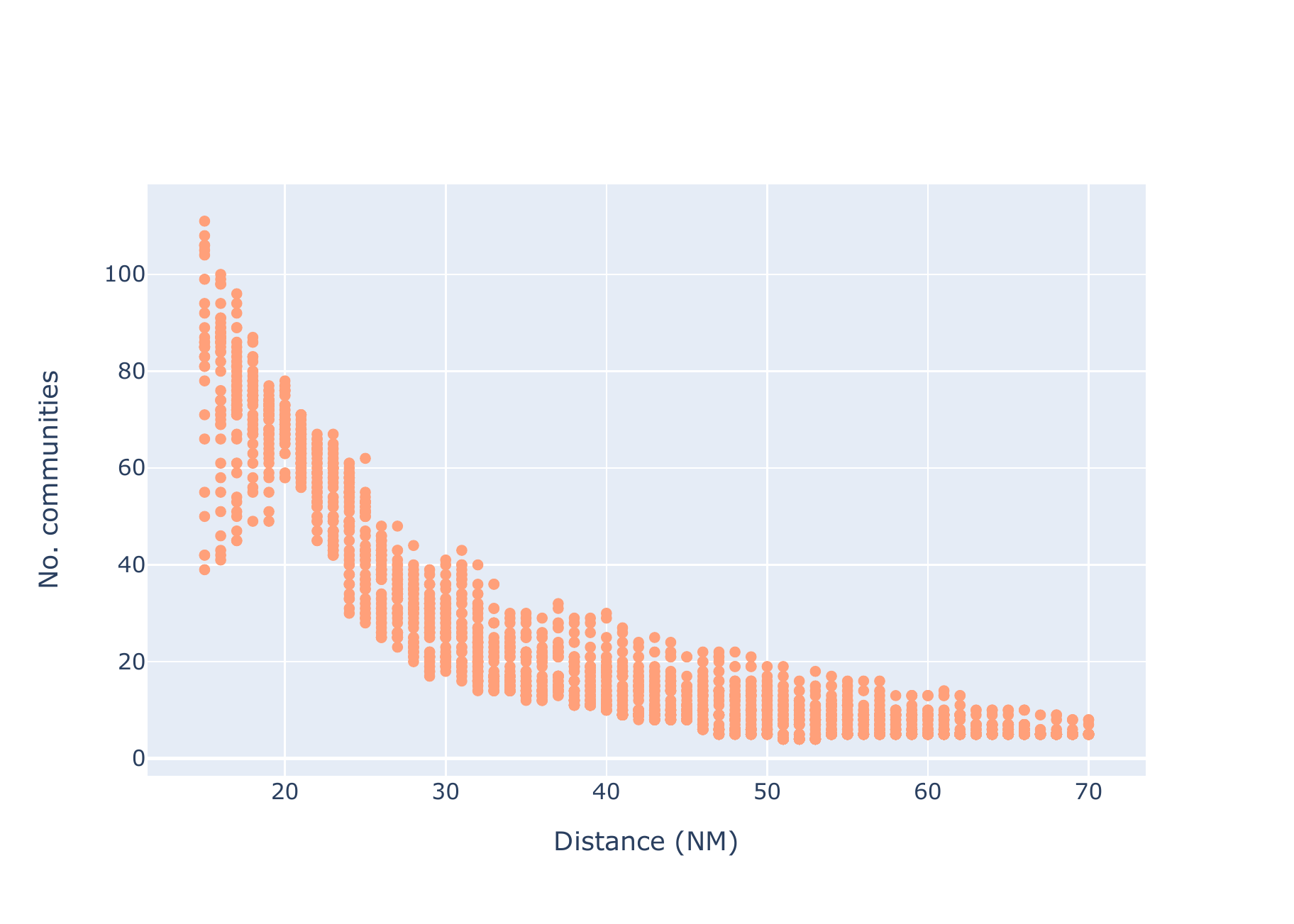}
\caption{Number of communities for different distance thresholds.}
\label{nocomm}
\end{minipage}
\hspace*{\fill} % it's important not to leave blank lines before and after this command
\begin{minipage}[t]{0.49\textwidth}
\includegraphics[width=1.0\linewidth,keepaspectratio=true]{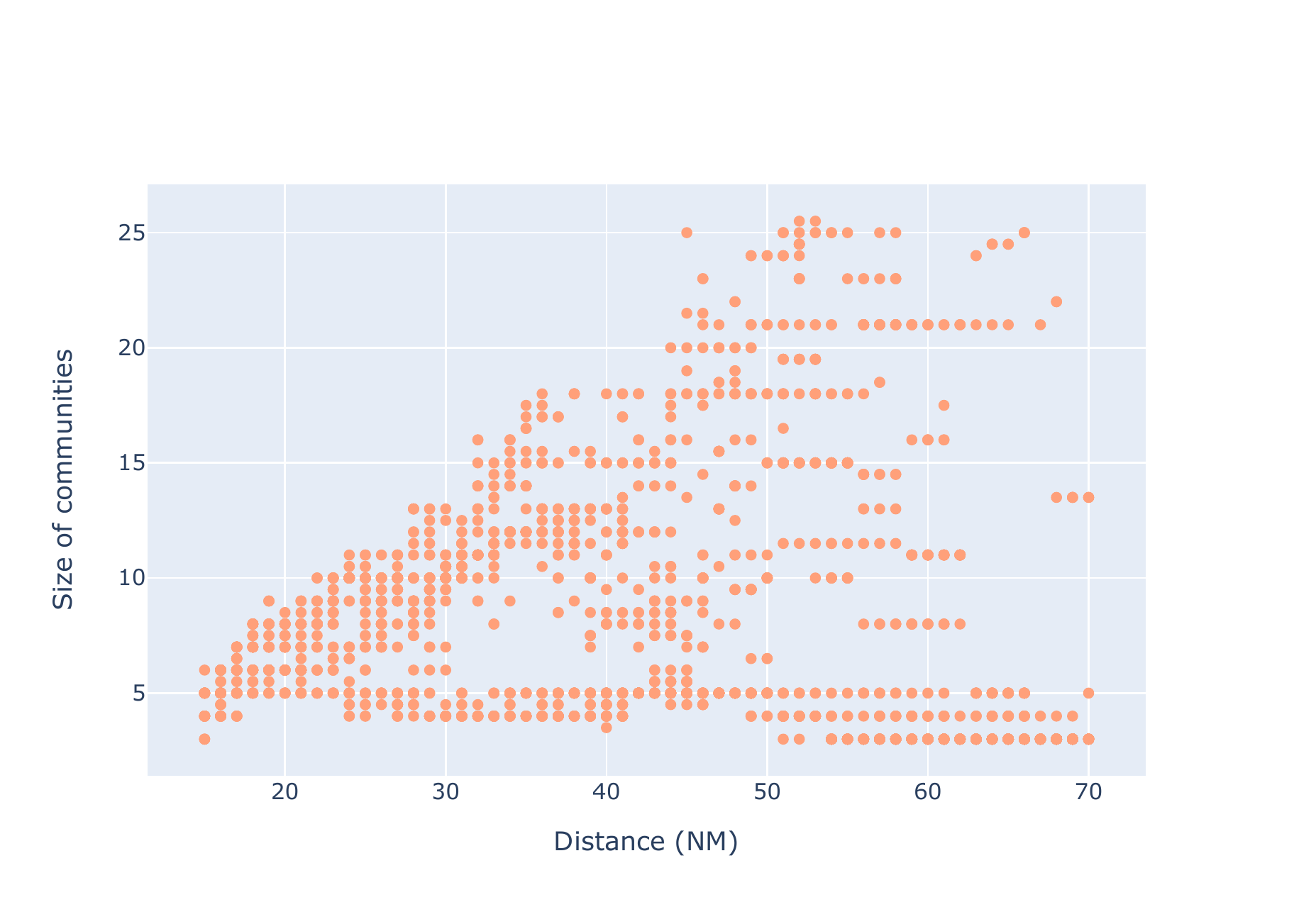}
\caption{Size of communities for different distance thresholds.}
\label{size}
\end{minipage}
\end{figure}

\begin{figure}[h!]
\centering
\includegraphics[width=0.8\linewidth,keepaspectratio=true]{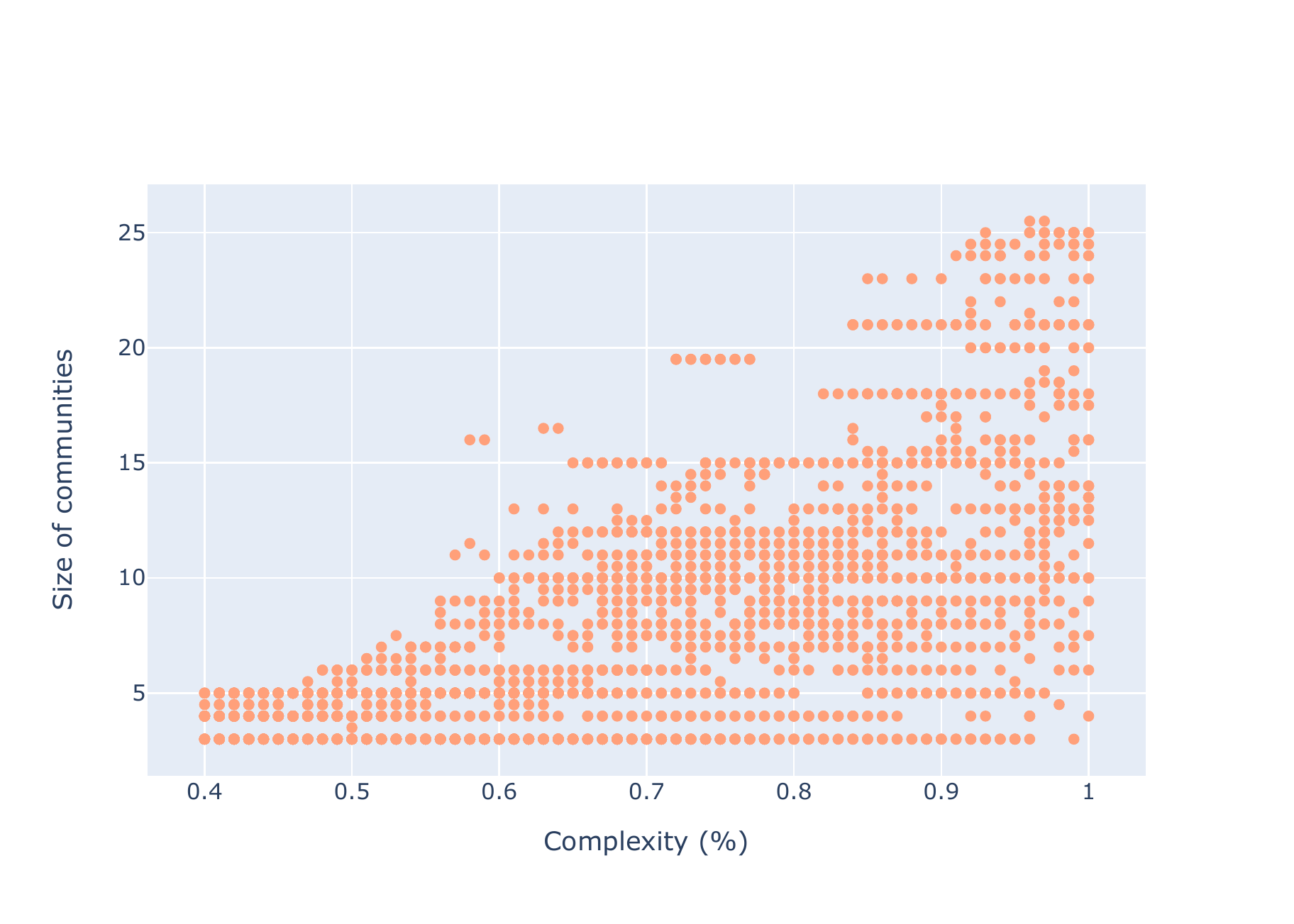}
\caption{Size of communities for different complexity thresholds.}
\label{size-comp}
\end{figure}

The results of the sensitivity analysis are shown in Figure \ref{fig:sobol}. As it can be observed, for all outputs, the total sensitivity Sobol indices indicate that the the maximal distance threshold for the interdependencies is the input that affects the output the most. In this work, the graph and in turn the communities are generated by considering the distance threshold in order to build the edges. Thus, it is evident that this should affect the number of communities and size of communities the most. For instance, when the distance is large enough, then at time $t$ it is reasonable to expect that the traffic graph is fully connected. In such a scenario, the complexity threshold would be irrelevant, as the community induced by the fully connected graph is always responsible for 100\% of the complexity at time $t$. A similar example could be used also to illustrate why this happens for the size of the communities. A large (or small) enough distance threshold will largely dictate which aircraft form interdependencies. In either extreme value, it is clear to see that the relevant aircraft would be forming the complex communities, thus determining also the size of the communities. Figures \ref{nocomm} and \ref{size} are further evidence of this. As it can be seen, the number and size of communities is strongly dependent on the distance threshold. The number of communities is inversely proportional with the threshold, while the size of the communities is proportional. We also observe a correlation between the size of communities with the complexity threshold, shown in Figure \ref{size-comp}. Nevertheless, as the sensitivity analysis suggests, this is as a result of the correlation between the two input parameters. In order for a community to be responsible for 100\% of the complexity, it should contain all aircraft that have at least one interdependency, which can be the case for bigger distance thresholds.

\begin{figure}[h!]
\centering
\includegraphics[width=0.8\linewidth,keepaspectratio=true]{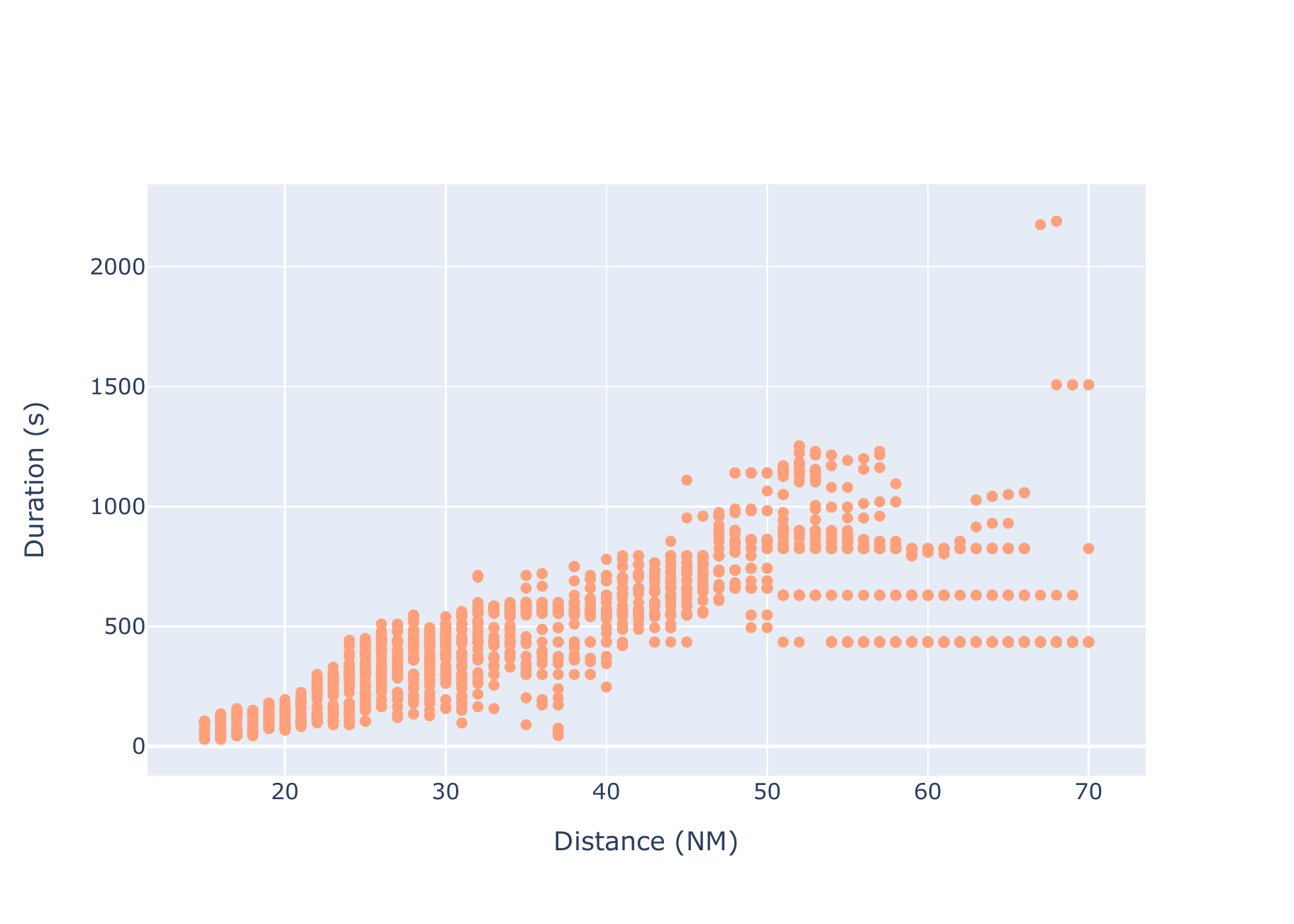}
\caption{Duration of communities for different distance thresholds.}
\label{duration-dist}
\end{figure}

From Figure \ref{fig:sobol}, it can be observed that the median duration is also affected mostly from the distance threshold, further evidenced in Figure \ref{duration-dist}. There, we can see that when the distance threshold is low, communities tend to last less. This happens because smaller distance thresholds are very sensitive towards the changes in the aircraft positions, as evidenced also from Figure \ref{nocomm}, where smaller distance thresholds lead to many present communities. 

\begin{figure}[h!]
\begin{minipage}[t]{0.49\textwidth}
\includegraphics[width=1.0\linewidth,keepaspectratio=true]{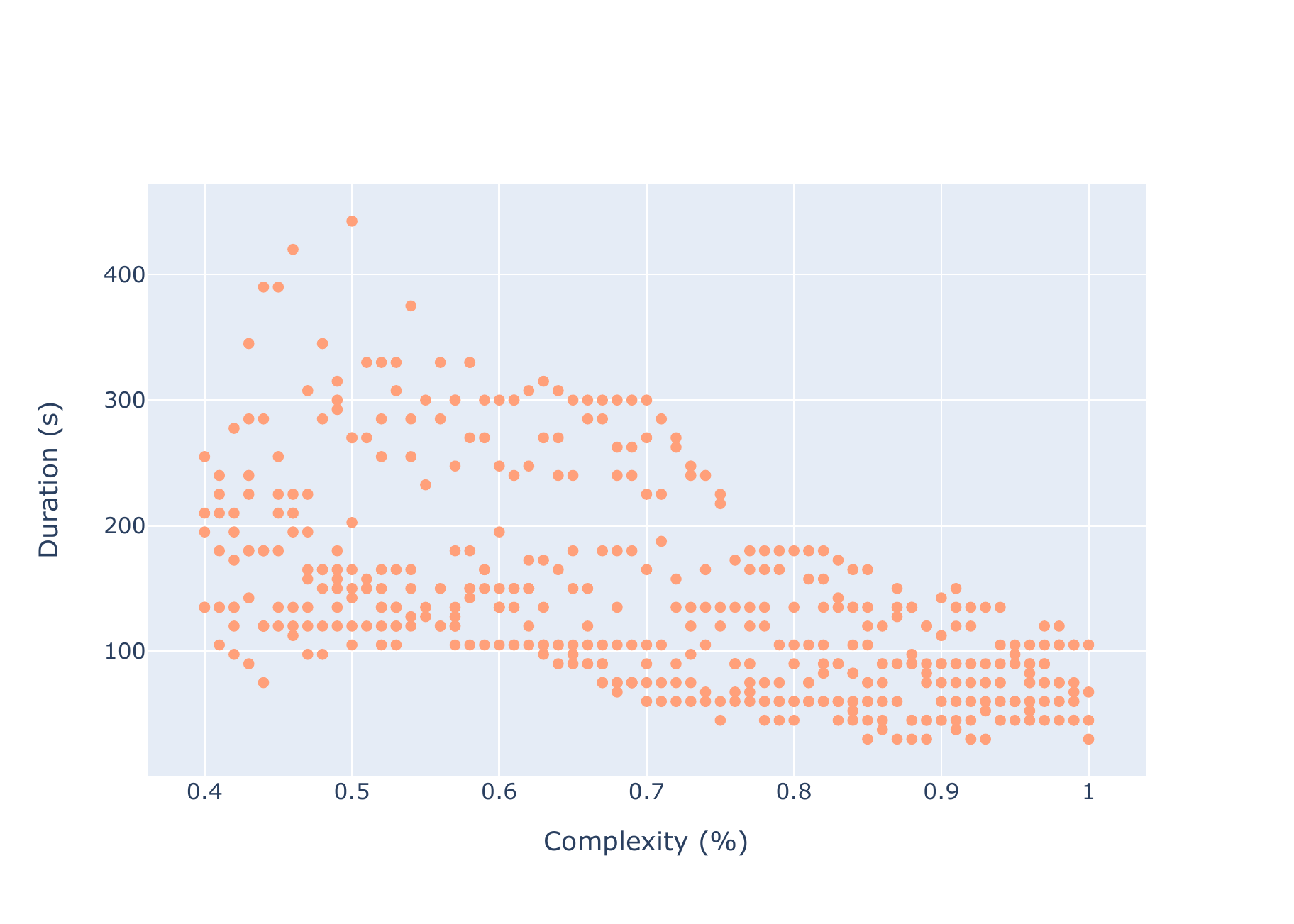}
\caption{Duration of communities for different complexity thresholds (Distance threshold < 25 NM).}
\label{u25}
\end{minipage}
\hspace*{\fill} % it's important not to leave blank lines before and after this command
\begin{minipage}[t]{0.49\textwidth}
\includegraphics[width=1.0\linewidth,keepaspectratio=true]{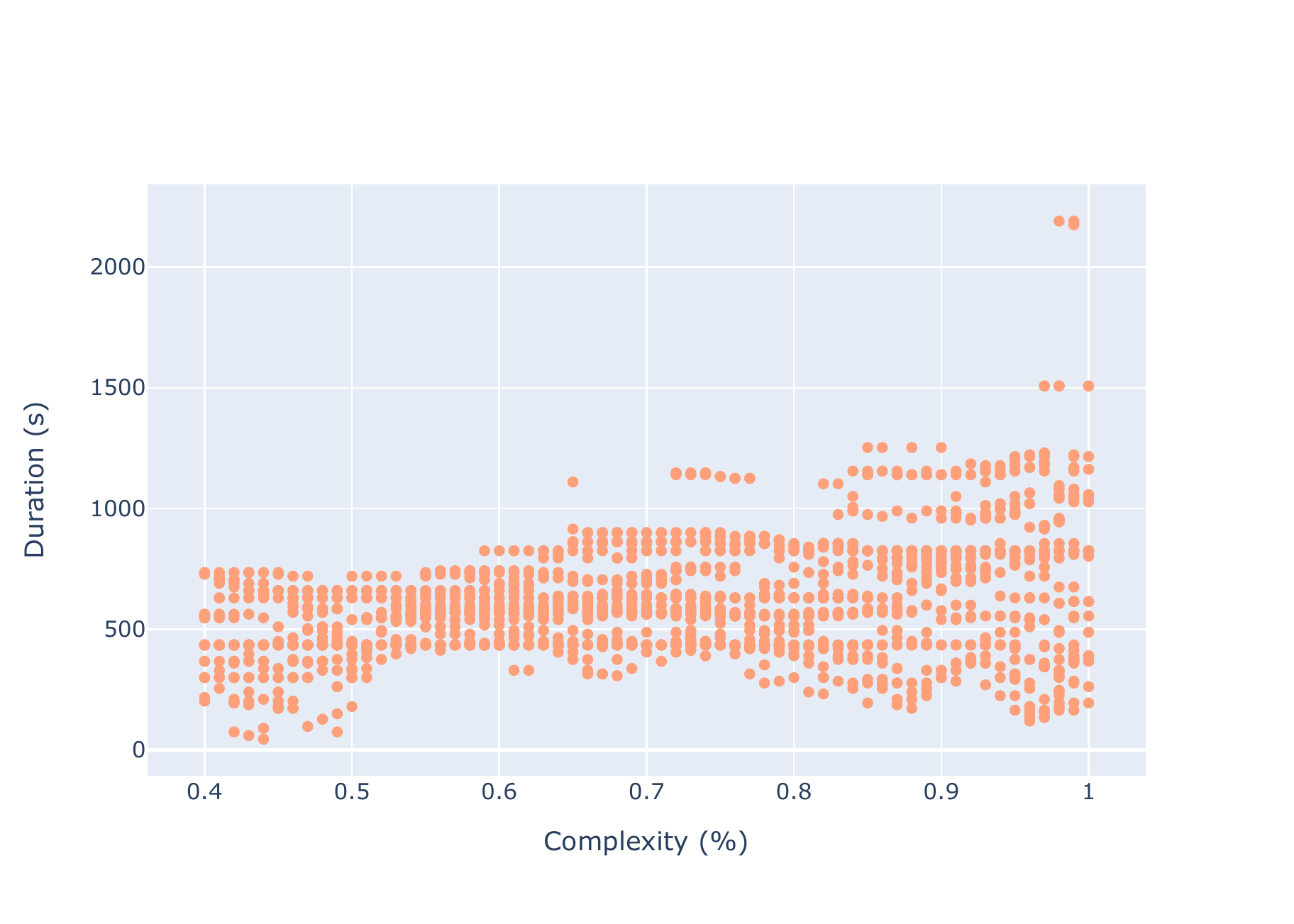}
\caption{Duration of communities for different complexity thresholds (Distance threshold > 25 NM).}
\label{o25}
\end{minipage}
\end{figure}

To give some insight on how the median duration is affected by the complexity threshold, we show Figures \ref{u25} and \ref{o25}. When the distance threshold is less than 25 NM, we observe that the duration decreases with increased complexity threshold. Such a result can be explained by the fact that with such a small distance threshold, it is more likely to have communities that are comprised of a subset of the aircraft co-existing in time in the sector. Consequently, aircraft can join communities and allow for communities to exist for longer. Nevertheless, this interpretation does not tell the whole story, as it can be observed that in such a setting the longest community existed for around 400s. On the other hand, when the distance threshold is bigger than 25 NM, the duration of communities is affected less by the complexity threshold. In this setting, we can observe that most communities last less than 17 minutes (less than 1000 s). The different response for different distance thresholds explains the Sobol index for the complexity threshold in Figure \ref{fig:sobol}.

While we show that the algorithm is mostly affected by the distance threshold, this does not indicate what input values different ATM subsystems should use. In fact, the analysis that we conducted in this section should serve for practitioners (e.g., NM, ATC etc.) to determine what values are more suitable for their use-case. For instance, a similar analysis could be used as a baseline to quantify controller preferences, whether that be in the topology of the graph the traffic induces, or any of the other outputs provided by the propsoed algorithm in this paper. Finally, while we do not expect the behavior of the algorithm to fundamentally change with other datasets, the analysis conducted in this section also heavily depends on the sector (or any other division of airspace) that is being investigated.

\section{Conclusions}
In this work, we propose a methodology that extends existing air traffic complexity indicators based on dynamic graphs to provide highly granular and nuanced information. As such, the concept of \textit{single aircraft complexity} is proposed which measures the individual contribution of each aircraft to the overall sector complexity. Furthermore, the algorithm provides \textit{complex communities}, which are connected components of the air traffic graph in order to determine complex spatio-temporal areas in the sector. 

To effectively illustrate the algorithm, a web application was developed which visualizes several outputs of the algorithm, namely: a complexity animation, a strength indicator animation, a heatmap of complex communities and a summary table of complex communities. Furthermore, the tool also provides the user with the possibility to download a summary file for the scenario being investigated. The tool (and the underlying algorithm) are envisioned as neutral aids that can ease the smooth functional transition between ATM layers and DSS tools that should be used in union with existing tools. Furthermore, the information provided could enhance equity, fairness at the aircraft of airline granularity level. 

In order to support our claims, we provide detailed use cases based on synthetic traffic, as well as real historical traffic. We first show that the algorithm can serve to formalize controller decisions as well as guide controllers to better decisions in situations where multiple pairwise conflicts co-exist in time. Further, we investigate how the provided information can be used to increase transparency of the decision makers towards different AUs, which serves also to increase fairness and equity. Moreover, the algorithm was evaluated using historical traffic, which was regulated through delays to several aircraft as a result of ATC capacity. We constructed two scenarios: one with the regulations applied and the other without the regulations applied and showed how the complex communities were affected in three 30 minute time windows. Finally, an extensive sensitivity analysis for two of the inputs to the algorithm (maximum distance threshold to form interdependencies and the complexity threshold for a community to be considered complex) was conducted. The sensitivity analysis was conducted for three outputs of the system: number of communities, median size of communities and median duration communities. We found that the maximum distance threshold affected these outputs the most. To fully understand this result, the response of each output to the different input values was studied. We argued how a similar analysis could be used to quantify controller preferences for graph topologies in the sector.

Nevertheless, the proposed algorithm should be extended and further refined. Most importantly, as one of the inputs to the tool are trajectories in time, a way to consider uncertainties should be investigated. As previously mentioned, the tool is envisioned to be used alongside existing tools, therefore one way to consider uncertainty would be for the inputs to the tool to have modelled uncertainty beforehand. However, it could be interesting to extend the definition of the graph to contain not a weight for any interdependency, but a distribution of weights. 

Moreover, the input parameters of the algorithm are envisioned to be tuned according to the problem and preferences of the practitioners. Nevertheless, the tool could have several "modes" when given a certain value for the distance and complexity thresholds. For instance, a more conservative mode could involve considering larger inputs that would lead to bigger communities that last longer, but visually provide more information about aircraft that are further in the community. Through the animation output, the practitioner could still visualize the core of the community. A less conservative mode would instead only consider the core of the communities, in order to provide only crucial members of the communities. 

The information provided should be evaluated by practitioners in a user study in order to optimize how and what information is shown. Finally, the tool should be further developed into a mature DSS, in order to be used alongside existing tools and methodologies. 

\section*{Acknowledgements}
We thank Xin Wang for the insights on the traffic data.
This work was supported in part by the H2020 EngageKTN grant agreement No 783287 and CRIDA scholarship to research, development and innovation in ATM.

\bibliographystyle{plain}
\bibliography{biblio.bib}

%Bibliography
%\bibliographystyle{unsrt}  
%\bibliography{bibliography.bib}  

\end{document}